\documentclass[10pt,twocolumn,letterpaper]{article}
\usepackage[T1]{fontenc}
\PassOptionsToPackage{table}{xcolor}
\usepackage[pagenumbers]{cvpr}
\usepackage{times,graphicx,amsmath,amssymb,booktabs,multirow}
\usepackage{microtype}
\definecolor{cvprblue}{rgb}{0.21,0.49,0.74}
\usepackage[breaklinks,colorlinks,allcolors=cvprblue,linktoc=all,bookmarksnumbered,bookmarksopen,bookmarksdepth=2]{hyperref}
\def\confName{CVPR}
\def\confYear{2027}
\newcommand{\method}{ReScene}
\title{ReScene: Incremental Scene Chronology from Sparse Revisits}
\author{
Wei Qin Chuah \quad Zeji Hui \quad Amirali Khodadadian Gostar \quad Alireza Bab-Hadiashar\\
RMIT University\\
{\tt\small \{wei.qin.chuah, zeji.hui, amirali.khodadadian, alireza.bab-hadiashar\}@rmit.edu.au}
}
\hypersetup{pdftitle={ReScene: Incremental Scene Chronology from Sparser Revisits},
 pdfauthor={Wei Qin Chuah, Zeji Hui, Amirali Khodadadian Gostar, Alireza Bab-Hadiashar}}
\makeatletter
\def\cvprsect#1{\cvprsection[#1]{\texorpdfstring{\hskip -1em.~}{}#1}}
\def\cvprsubsect#1{\cvprsubsection[#1]{\texorpdfstring{\hskip -1em.~}{}#1}}
\renewcommand*\l@section[2]{\addvspace{0.55em}\begingroup\bfseries\@dottedtocline{1}{0pt}{2.7em}{#1}{#2}\endgroup}
\makeatother
\begin{document}
\maketitle
\addtocontents{toc}{\protect\setcounter{tocdepth}{-1}}

\begin{abstract}
Scene chronology aims to reconstruct a scene across visits separated by hours or days, during which objects may change in position, presence or shape. Such reconstruction must retain every earlier state, yet each revisit often provides only a few images. Storing a complete model for each visit duplicates the unchanged scene, whereas updating a single shared model alters earlier states. We present ReScene, which stores each visit as an immutable compact update, so that any earlier visit can be recovered exactly by applying the stored updates in order, without retaining past images. To localise each update, ReScene attributes the disagreement between the preceding reconstruction and the new images to individual Gaussians through their compositing weights. This attribution guides the suppression of outdated Gaussians and the refinement of persistent ones, while newly visible surfaces are initialised from learned multi-view depth. We also introduce Multi-State, a benchmark of non-rigid deformation and partial material removal, including synthetic scenes with ground-truth geometry. ReScene achieves quality higher than sequential 3DGS checkpointing with 2.5--4.8$\times$ less uncompressed history storage. At comparable PSNR, it also reduces foreground depth error by 3.2$\times$ relative to sequential ChronoGS, showing that accurate rendering does not necessarily imply accurate geometry. Code and benchmark are available at \href{https://waychin-weiqin.github.io/ReScene/}{GitHub}. 

\end{abstract}
\begin{figure}[t]
\centering
\includegraphics[width=0.95\columnwidth]{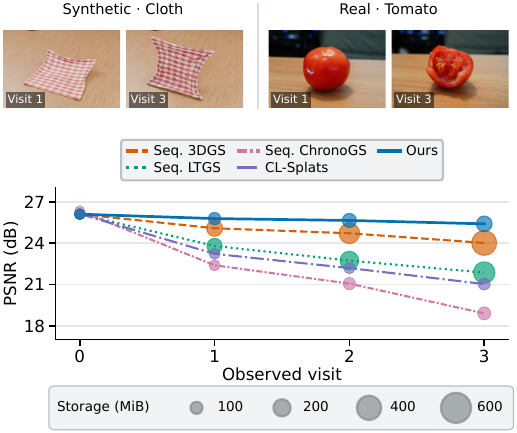}
\caption{\textbf{Sample data and history storage on Multi-State.}
\textit{Top:} deformation of a cloth (synthetic) and material removal from a tomato (real).
\textit{Bottom:} quality and history storage comparison on real Multi-State. Each point averages held-out PSNR over all visits observed so far, each recovered after the latest update. Marker area denotes history storage. \method{} retains the quality of earlier visits at a fraction of the storage required by checkpointing (Seq.~3DGS).}
\vspace{-13pt}
\label{fig:teaser}
\end{figure}

\vspace{-10pt}

\section{Introduction}
\label{sec:intro}
Revisiting a scene hours or days apart may reveal objects that change in position, presence or shape, while most of the scene remains unchanged. Retaining each of these states as a scene chronology (SC) supports retrospective inspection on construction sites or in warehouses, where the scene may need to be examined exactly as it was at an earlier visit. In practice, however, each revisit is often captured with only a few images, and earlier captures may no longer be available. A practical SC method must therefore update the reconstruction from sparse views without access to earlier images, and must never overwrite an earlier state, while keeping storage manageable as visits accumulate.

Existing methods meet only part of these requirements. For instance, dynamic methods such as 4DGS learn continuous motion from densely sampled video~\cite{4dgs,deformable}, whereas revisits observe distinct states without the transitions between them. SC methods jointly optimise all visits~\cite{neusc,chronogs,ltgs} and therefore require the images of every visit during training. A simple sequential alternative is to fine-tune the reconstruction at each visit and save a checkpoint, but this duplicates the unchanged scene, so storage grows redundantly with every visit. Another approach is to update a single shared model, which alters earlier states~\cite{gaussianupdate}. 

Restricting each update to the changed parts of the scene avoids both problems, but sparse revisits make this challenging. With few views, disagreement between the existing reconstruction and the new images may arise from either scene change or reconstruction error, and partial-object changes such as deformation confine it to small regions. Inaccurate localisation can therefore either prevent necessary updates or alter persistent surfaces. Moreover, newly exposed surfaces, such as the interior revealed by opening a grill lid, must  be initialised from these views alone. Recent continual methods update a reconstruction locally~\cite{clsplats,gaussianupdate}. GaussianUpdate matches instance masks and initialises new geometry from structure-from-motion points~\cite{gaussianupdate}, whereas CL-Splats detects change from semantic features and assigns it to Gaussians by voting with their projected centres~\cite{clsplats}. Such cues may miss changes confined to a small part of an object, and triangulation requires view overlap that sparse revisits may not provide~\cite{colmap}.

In this paper, we introduce \method{}, which localises updates to individual Gaussians and stores each visit as a compact update rather than a complete model. We attribute the photometric disagreement between the preceding reconstruction and the new images to Gaussians through their compositing weights, so that each Gaussian receives change and stable support in proportion to its contribution to the rendered pixels. This support bounds the opacity reduction of each Gaussian. Gaussians of moved or removed surfaces are therefore faded, whereas those that mainly render stable regions are largely protected. Rather than optimising an independent removal factor for each Gaussian~\cite{gaussianupdate}, a visit-specific network predicts suppression within this bound from Gaussian features, sharing it across Gaussians with similar features. Newly visible surfaces are initialised from learned multi-view depth that is consistent across the new views. Finally, each visit is stored as an immutable update of per-Gaussian opacities, selected attribute corrections and new Gaussians, so every earlier visit is recovered exactly without retaining past images or replaying earlier views. As shown in Fig.~\ref{fig:teaser}, \method{} approaches the quality of checkpointing (Seq. 3DGS) at a fraction of its storage.

Evaluating SC under such changes also exposes a gap in existing benchmarks, which mainly capture objects being moved, added, removed or articulated, and thus preserve the geometry of each object or rigid part. Non-rigid changes instead alter the shape of an object itself, either partly, as when a bite is taken out of a tomato, or wholly, as when a tissue is crumpled. Moreover, as ground-truth geometry is difficult to obtain for real scenes at every visit, existing benchmarks evaluate rendered appearance but not geometric accuracy, which matters in applications such as product damage monitoring. We therefore introduce \emph{Multi-State}, whose real scenes capture non-rigid changes under practical capture conditions and whose synthetic scenes provide ground-truth geometry at every visit (Fig.~\ref{fig:teaser}, top). On these scenes, we find that accurate rendering from held-out views does not necessarily imply accurate geometry.

Our contributions are summarised as follows:
\begin{itemize}
\item We present \method{}, which stores each visit as an immutable compact update, so every earlier visit can be recovered exactly without past images. Across Multi-State, WAT and ChronoScene, it achieves reconstruction quality comparable to or higher than that of 3DGS checkpointing with a 2.5--4.8$\times$ reduction in uncompressed history storage, whereas a sequentially updated ChronoGS loses 4.16--8.67\,dB on earlier visits.
\item We propose Gaussian-level change attribution through compositing weights, which bounds suppression to protect persistent surfaces, together with depth-seeded initialisation of newly visible surfaces. Compositing-weighted attribution improves PSNR by 2.70\,dB over projected-centre voting, and \method{} shows its largest advantage under the sparsest revisits.
\item We introduce Multi-State, to our knowledge the first SC benchmark with non-rigid deformation and partial material removal. Its real scenes capture changes such as a crushed can and a bitten tomato under practical capture conditions, whereas its synthetic scenes provide ground-truth geometry at every visit for evaluating geometric accuracy. 
\end{itemize}

\vspace{-5pt}
\section{Related Work}

\begin{figure*}[ht]
\centering\includegraphics[width=\textwidth]{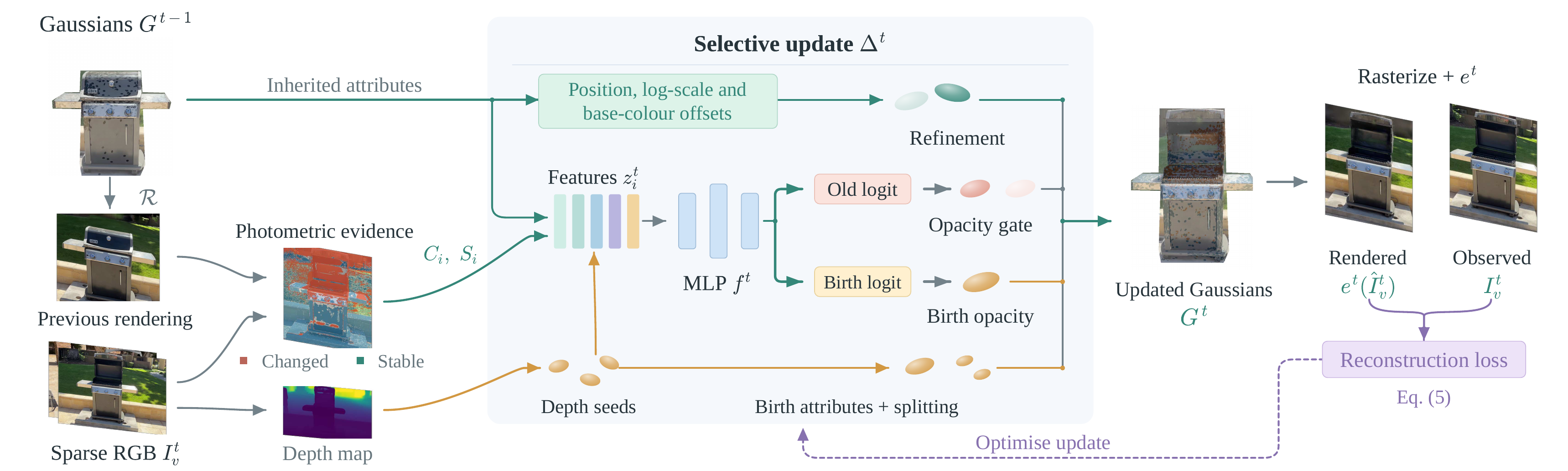}
\caption{\textbf{Overview of \method{}.} For each visit $t$, the preceding reconstruction $G^{t-1}$ is rendered from the viewpoint of the new images, and the disagreement between the renderings and the images is attributed to individual Gaussians as change support $C_i$ and stable support $S_i$. A visit-specific MLP $f^t$ maps the features $z_i^t$ of each Gaussian to opacity logits, which suppress Gaussians of moved or removed surfaces and determine the opacity of births. Gaussian offsets refine existing Gaussians, whereas births are initialised from depth seeds, and densified by splitting. The resulting update $\Delta^t$ produces $G^t$, whose rendering after the visit-specific appearance correction $e^t$ is compared with the observed images to optimise Eq.~\eqref{eq:objective}.}
\vspace{-10pt}
\label{fig:workflow}
\end{figure*}

\vspace{-5pt}
\paragraph{Scene reconstruction.}
NeRF~\cite{nerf} and 3D Gaussian Splatting (3DGS)~\cite{3dgs} reconstruct scenes for novel-view synthesis, and subsequent extensions accommodate appearance variation in unconstrained photo collections~\cite{nerfw,wildgaussians}. When only a few views are available, sparse-view approaches regularise optimisation with geometric priors~\cite{regnerf,fsgs,dngaussian} or predict representations directly from the input images~\cite{pixelsplat,mvsplat}. Learned geometric models such as DUSt3R~\cite{dust3r} and MASt3R~\cite{mast3r} further support reconstruction and initialisation from sparse observations~\cite{instantsplat}. Beyond static scenes, dynamic methods model motion through deformation fields~\cite{dnerf,nerfies,hypernerf,deformable,4dgs} or time-varying Gaussians~\cite{luiten2024dynamic,spacetimegs}. However, these methods reconstruct either a single scene configuration or motion within a continuous observation period. Scene chronology instead seeks to retain multiple historical scene states obtained during separate visits. New observations must be incorporated from only a few images, without overwriting previously recorded states.

\vspace{-12pt}
\paragraph{Scene chronology and continual reconstruction.}
Neural Scene Chronology~\cite{neusc} models temporal changes in scene content, ChronoGS~\cite{chronogs} separates persistent and changing components through temporally modulated anchors, and LTGS~\cite{ltgs} supports sparse-view updates by sharing object templates across visits. These methods jointly optimise the observations of all visits, whereas continual approaches incorporate new captures sequentially. Among the latter, CLNeRF~\cite{clnerf} relies on generative replay, CL-NeRF~\cite{clnerfwu} combines expert adaptation with knowledge distillation, and CL-Splats~\cite{clsplats} performs local updates with reversible history records. GaussianUpdate~\cite{gaussianupdate} is closely related to our work, as both update an existing Gaussian reconstruction from new observations. Its pipeline uses instance-mask correspondence to identify persistent content and COLMAP-derived points to initialise emerging geometry. Our setting explicitly targets sparse revisits with non-rigid deformation affecting either part or all of an object, where instance correspondence alone may not identify persistent surfaces and limited view overlap may constrain triangulation. We therefore attribute change and stable support to individual Gaussians and initialise new surfaces from learned multi-view depth.

\vspace{-5pt}

\section{Method}
\label{sec:method}
Given an initial 3DGS~ reconstruction $G^0$ built from the images of the first visit ($t=0$), each revisit $t=1,\dots,T$ provides a limited set of calibrated images $\mathcal{D}^t=\ (I^t_v,\Pi^t_v)\}_{v=1}^{n^t}$, where $\Pi^t_v$ denotes the camera parameters of image $I^t_v$ in a shared world frame. The scene is static within each visit but may change between visits. At visit $t$, our goal is to update $G^{t-1}$ to a reconstruction $G^t$ of the current scene while keeping every earlier visit exactly recoverable. Crucially, only $\mathcal{D}^t$ is available at this visit, as earlier images are not retained. Moreover, each revisit captures far fewer views than the first, so much of the scene may not be observed. As these regions provide no evidence of change, the update retains them as last reconstructed, and a change in such a region is incorporated only once a later visit observes it. 

As discussed in Sec.~\ref{sec:intro}, restricting each update to the changed regions of the scene avoids both the redundant storage of checkpointing and the alteration of earlier states caused by updating a single shared model. Under sparse revisits, however, this restriction requires distinguishing scene change from reconstruction error with limited observations and initialising newly visible surfaces from few views, both of which are challenging.


To address these challenges, ReScene learns a compact update $\Delta^t$ to $G^{t-1}$ that stores only what changes, consisting of Gaussian offsets and a visit-specific appearance correction $e^t$ that absorbs global changes such as exposure. To distinguish change from reconstruction error with few observations, ReScene weighs change support against that of stability for each Gaussian before suppressing or refining it. To initialise newly visible surfaces from these few views, ReScene uses learned multi-view depth that is consistent across the new views, rather than triangulation, which  requires greater view overlap. Any earlier state is recovered as
\begin{equation}
G^\tau = G^0 \oplus \Delta^1 \oplus \cdots \oplus \Delta^\tau, \quad \tau \le t,
\end{equation}
and rendered with $e^\tau$, where $\oplus$ applies an update and $e^0$ is the
identity (Sec.~\ref{sec:history}). We denote the differentiable rasterisation of a
state $G$ from camera $\Pi^t_v$ by $\mathcal{R}(G,\Pi^t_v)$. As summarised in
Fig.~\ref{fig:workflow}, each update is obtained by gathering evidence from the
new visit (Sec.~\ref{sec:evidence}), selectively updating Gaussians
(Sec.~\ref{sec:updates}) and optimising these updates jointly
(Sec.~\ref{sec:optimisation}).

\subsection{Evidence from the New Visit}
\label{sec:evidence}
Before updating the model, we gather evidence of disagreement and stability from the new observations, which guides suppression, refinement and the placement of new Gaussians.
\vspace{-20pt}
\paragraph{Pixel-level disagreement.}
If a region of the scene remains unchanged at the subsequent visit, the model updated at the previous visit should reproduce its appearance from the new viewpoints. We therefore compare each new image $I_v^t$ with the corresponding rendering $\mathcal R(G^{t-1},\Pi_v^t)$ using the pixelwise absolute photometric error. However, global colour shifts between visits (e.g. changes in exposure) could increase the photometric error across the entire image. To handle this, we first subtract a per-channel colour offset from the image. The offset is estimated only from low-error pixels so that locally changed regions do not bias it. After this correction, we compute the change mask $M_v^{c,t}$ by filtering and thresholding the recomputed error, and then removing small connected components to suppress noise. Low-error pixels outside this mask form the stable mask $M_v^{s,t}$, whereas the remaining ambiguous pixels are unassigned (Fig.~\ref{fig:workflow}). As inconsistencies may also arise from reconstruction errors rather than physical changes, we only consider these masks as evidence of change.
\vspace{-11pt}
\paragraph{Gaussian-level support.}
The change and stable masks are defined over pixels, whereas the model is updated at the level of individual Gaussians. As each rendered pixel is formed by alpha compositing the Gaussians along its ray, we transfer the evidence of each pixel to these Gaussians using their compositing weights. The compositing weight $a_{iv}^t(p)$ of Gaussian $i$ at pixel $p$ in view $v$ combines the opacity of the Gaussian, its projected footprint and the transmittance of the Gaussians in front of it, and therefore measures how much Gaussian $i$ contributes to the colour of that pixel. Accumulating the masks with these weights over all views of the current visit gives the change support $C_i^t$ and the stable support $S_i^t$ of each Gaussian:
\begin{equation}
 \begin{aligned}
 C_i^t&=\frac{1}{n^t}\sum_{v,p}a_{iv}^t(p)M_v^{c,t}(p),\\
 S_i^t&=\frac{1}{n^t}\sum_{v,p}a_{iv}^t(p)M_v^{s,t}(p).
 \end{aligned}
 \label{eq:contribution-evidence}
\end{equation}
Consequently, occluded Gaussians receive little support, and ambiguous pixels, which belong to neither mask, contribute to neither of the support terms. Both terms are computed efficiently during the backward pass of the renderer, given that the derivative of a pixel's colour with respect to a Gaussian's colour equals its compositing weight. This accumulates the masked contributions for each Gaussian without explicitly storing all Gaussian--pixel weights. 

However, the magnitude of $C_i^t$ depends not only on change but also on the overall contribution of a Gaussian to the renderings. For example, a Gaussian whose projected footprint spans many pixels, or is visible in many views, accumulates more support than a Gaussian with a smaller footprint or fewer observations. Consequently, $C_i^t$ alone does not indicate whether the contribution of the Gaussian lies predominantly in changed or stable regions. We therefore normalise the change support by the sum of the change and stable supports to obtain the change weight
\begin{equation}
 w_i^t=\frac{C_i^t}{C_i^t+S_i^t+\epsilon},
 \label{eq:change-weight}
\end{equation}
which is the fraction of its accumulated support that lies in changed pixels, where the small constant $\epsilon$ prevents division by zero. The masks and support are computed once from $G^{t-1}$ and remain fixed during optimisation, so that the change weights do not drift as the model is updated.
\vspace{-10pt}
\paragraph{Depth support.}
The change masks localise disagreement in the image plane but do not determine where new surfaces lie in 3D. To this end, we estimate a depth map and confidence scores for each view of the current visit using MASt3R~\cite{mast3r}, with the cameras fixed to $\Pi_v^t$ to align the estimates with the reconstruction. Retaining only confident depth estimates, we assess geometric support through two comparisons. First, we back-project estimated depths and reproject the resulting points into other views, checking their cross-view consistency with the corresponding depth estimates. Second, we compare estimated depths with depths rendered from $G^{t-1}$ at the same cameras. Cross-view consistency supports the placement of new Gaussians, while agreement with the preceding reconstruction supports refining existing ones. 

\subsection{Selective Gaussian Updates}
\label{sec:updates}
Guided by the Gaussian-level support and depth support, we update $G^{t-1}$ by suppressing outdated Gaussians, refining persistent ones and adding Gaussians for new surfaces. Gaussian attributes outside the selected update sets remain unchanged.
\vspace{-12pt}
\paragraph{Suppressing outdated Gaussians.}
Gaussians associated with moved or removed surfaces may occlude content that becomes newly visible. To address this, we reduce their opacity while retaining their previous states in the update history, allowing earlier scene states to be recovered. The opacity of each Gaussian is updated as
\begin{equation}
 \alpha_i^t=\alpha_i^{t-1}
 \left[
 1-w_i^t
 \left(1-\sigma\!\left(f_{\mathrm{old}}^t(z_i^t)\right)\right)
 \right],
 \label{eq:gate}
\end{equation}
where $w_i^t$ is the change weight from Eq.~\eqref{eq:change-weight}, $\sigma$ denotes the sigmoid function and $f_{\mathrm{old}}^t$ is the suppression logit predicted by a visit-specific MLP (Fig.~\ref{fig:workflow}). The input $z_i^t$ includes the encoded position, base colour, image-based support and a birth indicator. As the opacity multiplier lies between $1-w_i^t$ and $1$, the change weight $w_i^t$ bounds the extent of suppression. For instance, Gaussians with $w_i^t=0$ retain their opacity, and Gaussians that contribute predominantly to stable regions can be suppressed only slightly. Within this bound, the MLP learns the opacity reduction of each Gaussian. The shared MLP predicts suppression from Gaussian features rather than optimising an independent parameter for each Gaussian, sharing the suppression function across Gaussians with similar features.


\vspace{-12pt}
\paragraph{Adding Gaussians for new surfaces.}
\label{sec:births}
Surfaces that appear between the two visits, or that become visible once an occluding surface is removed, are not represented in the model and require new Gaussians, which we refer to as \emph{births}. Births are placed by back-projecting changed pixels whose estimated depth is confident, consistent across at least two views of the current visit and different from the depth rendered from $G^{t-1}$, which reduces duplication of existing surfaces. Differences in either direction are accepted, as an added surface may lie in front of the previous geometry, whereas a removal may reveal a surface behind it. To limit the number of parameters optimised from sparse views, we keep the higher-order colour coefficients at zero. The opacity of birth $i$ is defined as \(\alpha_i^t=\sigma\bigl(\beta_i+f_{\mathrm{new}}^t(z_i^t)\bigr)\)
where $\beta_i$ is a learned per-birth logit and $f_{\mathrm{new}}^t$ is the birth opacity logit predicted by the same visit-specific MLP as $f_{\mathrm{old}}^t$ in Eq.~\eqref{eq:gate}. The parameter $\beta$ allows the opacity of each birth to be adjusted individually, while $f_{\mathrm{new}}^t$ adds a feature-dependent contribution shared across births. Births are ranked by accumulated positional gradient norm, and those with the largest values are split using a strategy inspired by the densification of 3DGS~\cite{3dgs}. The total number of births is limited to bound the storage added at each visit. The supplementary material reports the birth-limit study and the placement thresholds and splitting settings.


\vspace{-13pt}
\paragraph{Refining existing Gaussians.}
Surfaces represented in $G^{t-1}$ may need small corrections to their position or colour, which suppression cannot provide, and births would duplicate existing surfaces. To make these corrections, we refine a set $\mathcal F^t$ of Gaussians from $G^{t-1}$, comprising those that receive change or stable support and the conservative depth-supported subset described in the previous section. Details of the selection criteria for $\mathcal F^t$ are provided in the supplementary material. 
The depth-supported Gaussians lie where confident depth estimates from the current views agree with the depth rendered from $G^{t-1}$, which indicates that they already represent the observed surface. We apply additional photometric supervision to them through a soft refinement mask $M_v^{r,t}$. At each pixel, the mask sums their compositing weights, clips the sum to $[0,1]$ and multiplies it by the depth-agreement map, so that supervision is concentrated where these Gaussians are visible and their depth is confirmed. Regions identified as changed during depth-proposal generation are excluded, as their content is expected to be replaced rather than corrected. The remaining Gaussians in $\mathcal F^t$ are supervised through the change and stable masks, where we learn offsets to their corresponding centre, log-scale and view-independent colour coefficients. Centre offsets are expressed relative to the Gaussian's scale. Rotations and higher-order colour coefficients remain fixed to limit the attributes optimised from sparse observations and stored at each visit.
\vspace{-12pt}
\paragraph{Visit-specific appearance.}
Global appearance can vary between visits, for example due to changes in exposure. Rather than recolouring every Gaussian to absorb such variation, we learn a per-channel log-gain $a^t\in\mathbb R^3$ and bias $b^t\in\mathbb R^3$ for each visit, and apply the correction $e^t(X)=\exp(a^t)\odot X+b^t$ to the rendered images $X$ of that visit. The log-gain and bias are both initialised to zero, so that the correction is initially the identity, and are optimised jointly with the Gaussian updates.

\subsection{Optimisation}
\label{sec:optimisation}
The suppression network, births, refinement offsets and appearance correction, which together form the update $\Delta^t$, are optimised jointly using only the images of the current visit. As all quantities in this section refer to visit $t$, we omit the superscript $t$ except on the models. The objective serves two goals: to explain the new observations where the scene has changed, and to avoid altering the model where it has not.

For the first goal, we compare the corrected rendering $e(\hat I_v)$ with the observed image $I_v$, where $\hat I_v=\mathcal R(G^t,\Pi_v)$. The photometric loss $\mathcal L_{\mathrm{photo},v}^{k}$, which combines L$_1$ and D-SSIM terms as in 3DGS~\cite{3dgs}, is evaluated separately within the change, stable and refinement masks $M_v^{k}$, $k\in\{c,s,r\}$, and normalised by the total weight of each mask, so that a small changed region is not outweighed by the much larger stable region. For the second goal, a preservation term $\mathcal L_{\mathrm{preserve},v}$ penalises the absolute colour difference between the renderings of $G^t$ and $G^{t-1}$ at camera $\Pi_v$, both corrected by $e$, weighted by $M_v^{s}(1-M_v^{r})$. Unlike the stable photometric loss, which matches the new images, this term discourages unnecessary deviation from $G^{t-1}$ while still allowing refinement in depth-supported regions. The overall reconstruction objective is defined as
\begin{equation}
 \mathcal L=
 \frac{1}{N_v}\sum_v
 \Bigl[
 \sum_{k\in\{c,s,r\}}\mathcal L_{\mathrm{photo},v}^{k}
 +\mathcal L_{\mathrm{preserve},v}
 \Bigr],
 \label{eq:objective}
\end{equation}
where $N_v$ is the number of views. Only the current update is optimised, so the reconstructions of earlier visits remain unchanged. We provide the optimisation settings in the supplementary material.
\subsection{Compact History and Retrieval}
\label{sec:history}
After optimisation, each update $\Delta^t$ stores the opacity vector of existing Gaussians, the refined attributes and their indices, the attributes of births and the appearance correction $e^t$. Unmodified geometric and colour attributes are reused from preceding states, which reduces storage compared with saving a complete model after every visit (Sec.~\ref{sec:main-results}). Before saving, negligible opacities of existing Gaussians are set to zero without changing their indices, and births with low opacity are discarded. As the update stores explicit attribute values, decoding requires no network evaluation. Our implementation also retains the MLP weights and auxiliary arrays, which are included in all reported storage measurements. To recover visit $\tau$, we apply $\Delta^1,\ldots,\Delta^\tau$ sequentially to the initial reconstruction $G^0$ and render with $e^\tau$. Later updates leave these updates unchanged, so the stored reconstruction of each visit can be recovered exactly. This incremental representation also accommodates future visits, where each new visit is incorporated by optimising an additional update from its own images, without accessing earlier images or modifying previously stored updates.

\begin{table*}[!t]\centering\footnotesize
\setlength{\tabcolsep}{3pt}
\caption{\textbf{Reconstruction quality and history storage across three benchmarks.} Quality is reported as PSNR (dB) and SSIM (higher is better for both), averaged equally across scenes. Seq.~ChronoGS queries its latest shared model, so its arrival and final-recall quality are reported separately; the other methods preserve arrival quality through stored histories. Storage is the mean uncompressed model and history size per scene (MiB, lower is better). Bold marks the best final-recall quality and the lowest storage in each column.}
\vspace{-5pt}
\label{tab:overview}\label{tab:wat}\label{tab:chrono}\label{tab:multistate}
\begin{tabular*}{\textwidth}{@{\extracolsep{\fill}}lcccccccccccc@{}}\toprule
 & \multicolumn{3}{c}{Multi-State (Real)} & \multicolumn{3}{c}{Multi-State (Synthetic)} & \multicolumn{3}{c}{WAT} & \multicolumn{3}{c}{ChronoScene}\\
\cmidrule(lr){2-4}\cmidrule(lr){5-7}\cmidrule(lr){8-10}\cmidrule(l){11-13}
Method & PSNR & SSIM & Storage & PSNR & SSIM & Storage & PSNR & SSIM & Storage & PSNR & SSIM & Storage\\\midrule
CL-Splats & 21.03 & 0.826 & \textbf{91.2} & 29.34 & 0.942 & 220.5 & 18.10 & 0.731 & \textbf{174.7} & 18.74 & 0.605 & 1581.5\\
Seq. ChronoGS (arrival) & 24.92 & 0.863 & \multirow{2}{*}{108.6} & 32.07 & 0.954 & \multirow{2}{*}{\textbf{108.7}} & 27.42 & 0.891 & \multirow{2}{*}{368.0} & 20.69 & 0.613 & \multirow{2}{*}{\textbf{706.7}}\\
Seq. ChronoGS (recall) & 18.90 & 0.774 & & 25.81 & 0.907 & & 18.75 & 0.726 & & 16.53 & 0.467 & \\
Seq. LTGS & 21.87 & 0.844 & 304.7 & 28.81 & 0.938 & 995.0 & 23.24 & 0.834 & 873.2 & 19.41 & 0.617 & 5664.1\\
Seq. 3DGS & 24.02 & 0.861 & 410.5 & \textbf{33.24} & 0.956 & 1673.3 & \textbf{25.67} & \textbf{0.863} & 1816.1 & 19.57 & 0.577 & 9618.7\\
\midrule
Ours & \textbf{25.40} & \textbf{0.873} & 163.6 & 32.71 & \textbf{0.958} & 352.2 & 25.59 & 0.861 & 375.3 & \textbf{21.09} & \textbf{0.642} & 2387.4\\
\bottomrule
\end{tabular*}
\vspace{-12pt}
\end{table*}

\section{Experiments}
\label{sec:experiments}
\begin{table}[t]
\centering\footnotesize
\setlength{\tabcolsep}{3.8pt}
\caption{\textbf{Geometric error of foreground objects on synthetic Multi-State.} We report the mean absolute error (lower is better) of the rendered expected depth along the camera axis, evaluated using ground-truth foreground masks and rendered opacity $\geq0.5$. The reported error is averaged over all visits, including the initial one. The lowest error in each column is shown in bold.}
\vspace{-5pt}
\label{tab:hard-geometry}
\begin{tabular}{lccccc}\toprule
Method & Robot & Teddy & Plant & Cloth & Mean\\\midrule
Ours & \textbf{30.22} & \textbf{7.10} & 121.80 & \textbf{19.23} & \textbf{44.59}\\
Seq.~LTGS & 197.63 & 29.90 & \textbf{101.21} & 132.54 & 115.32\\
Seq.~ChronoGS (arrival) & 153.10 & 27.54 & 144.45 & 251.67 & 144.19\\
Seq.~ChronoGS (recall) & 1119.22 & 46.48 & 232.49 & 247.18 & 411.34\\
\bottomrule
\end{tabular}
\vspace{-10pt}
\end{table}
\subsection{Experimental Setup}
\label{sec:setup}
\vspace{-3pt}
\paragraph{Datasets.}
We evaluate on our Multi-State benchmark, which contains both rigid and non-rigid changes, including partial-object changes rarely found in existing benchmarks, such as a tissue being pulled out and crumpled or a bite taken out of a tomato. We evaluate four registered real scenes (16 visits) and four synthetic scenes (42 visits) from Multi-State. Its synthetic scenes provide ground-truth geometry at every visit, which enables geometry evaluation, whereas its real scenes contain the imperfections of real-world capture, such as sensor noise and camera poses that are estimated offline and aligned across visits. The construction of both parts is described in the supplementary material. We further evaluate on World-Across-Time (WAT)~\cite{clnerf}, which contains real revisits of the same scenes, and ChronoScene~\cite{chronogs}, which contains larger environments.

\begin{figure*}[!th]
\centering\includegraphics[width=\textwidth]{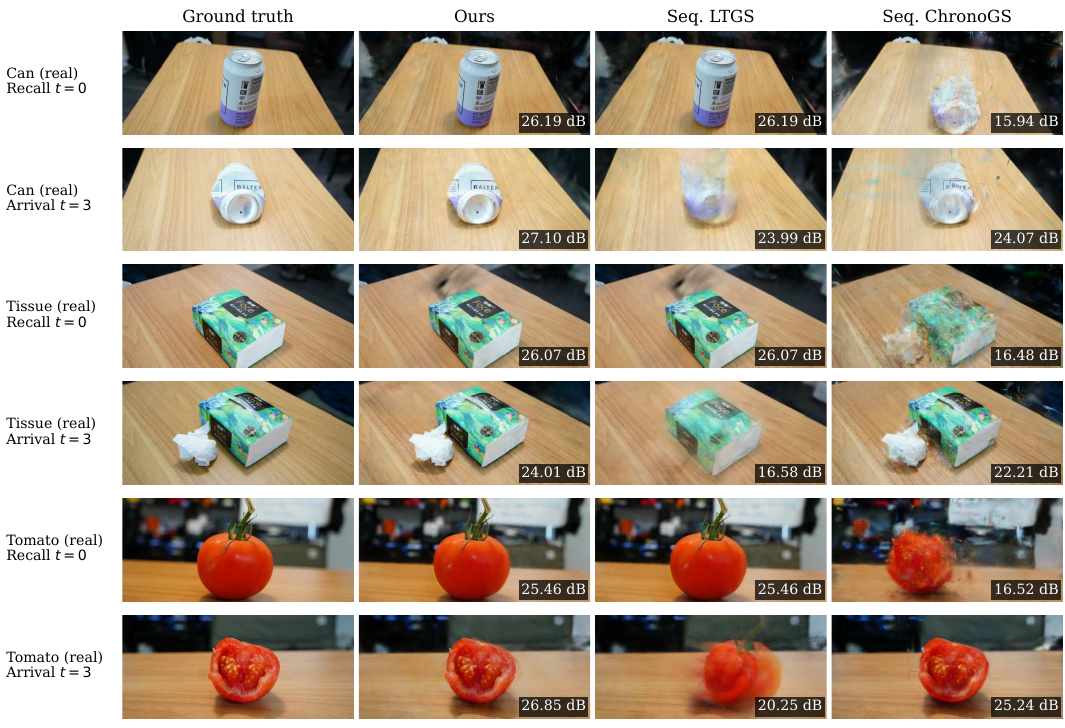}
\caption{\textbf{Final recall and arrival on real Multi-State.} Each scene pairs an earlier visit recovered after the final update (upper row) with a later visit at arrival (lower row). Our method and Seq.~LTGS recover earlier states exactly; Seq.~ChronoGS alters them. Cameras are matched within rows; overlays report whole-image PSNR.}
\vspace{-12pt}
\label{fig:wat}\label{fig:cross-dataset}\label{fig:paired-qualitative}
\end{figure*}

\vspace{-13pt}
\paragraph{Baselines.}
We compare with sequential 3DGS, LTGS~\cite{ltgs}, ChronoGS~\cite{chronogs} and CL-Splats~\cite{clsplats}. Sequential 3DGS fine-tunes all Gaussian attributes from the preceding model using current images, with densification and pruning. LTGS supports sparse-view updates through shared object templates and multi-period optimisation, whereas ChronoGS jointly represents visits using temporally modulated anchors. We adapt both to sequential access: Seq.~LTGS uses current images and synthetic replay from its preceding model, while Seq.~ChronoGS updates its shared representation using current images. Neither adaptation uses earlier or future captured images. Sequential 3DGS and our LTGS adaptation save complete models for historical retrieval; Seq.~ChronoGS queries its final shared model. CL-Splats locally updates its reconstruction and recovers earlier visits from reversible records. All methods share camera registration, selected training views and held-out cameras; differences in initialisation, optimisation and replay are detailed in the supplement. GaussianUpdate~\cite{gaussianupdate} is closely related, but the official implementation was unavailable publicly. The supplement compares our paper-guided WAT experiment with its published results and discloses the remaining protocol differences.

\vspace{-13pt}

\paragraph{Evaluation protocol.}
On every benchmark, the initial reconstruction is trained on all training images of the first visit, after which each later visit provides up to 24 images, without access to the images of earlier visits. We measure reconstruction quality at \emph{arrival} and at \emph{final recall}. At arrival, each visit is evaluated immediately after its reconstruction is obtained, using the held-out images of that visit. At final recall, each method recovers every observed visit from its retained representation after all sequential updates are complete. The rendered images are compared with the same held-out images used at arrival. As both evaluations use identical held-out images, a decrease from arrival to final recall indicates that subsequent updates have degraded earlier reconstructions. We report PSNR and SSIM alongside the total storage required to recover all visits. On the synthetic scenes of Multi-State, we additionally evaluate geometric accuracy as the error of rendered expected foreground depth with respect to the ground truth. All scene-wise results are provided in the supplementary document. 
\subsection{Main Results}
\label{sec:main-results}\label{sec:geometry}
\vspace{-6pt}
\paragraph{Quality and storage.}
Table~\ref{tab:overview} reports the reconstruction quality and the storage required to retain all visits on Multi-State, WAT and ChronoScene. Across all benchmarks, our method preserves the quality of every visit while requiring only a fraction of the storage needed for checkpointing. On the real scenes of Multi-State and on ChronoScene, our method achieves the highest PSNR and SSIM among all methods, exceeding Seq.~3DGS by 1.39 and 1.52\,dB, respectively, with storage reductions of 2.5$\times$ and 4.0$\times$. On the synthetic scenes of Multi-State and on WAT, our method remains within 0.53 and 0.08\,dB of Seq.~3DGS in PSNR, respectively, with storage reductions of 4.7$\times$ and 4.8$\times$. The Seq.~3DGS method stores a complete model for every visit, whereas our method stores selective updates to reduce duplication of unchanged attributes. Relative to Seq.~LTGS, which also stores complete checkpoints, our method improves PSNR by 1.68 to 3.90\,dB with storage reductions of 1.9$\times$ to 2.8$\times$. CL-Splats requires less storage than our method and preserves its arrival quality, but its reconstruction quality remains 2.35 to 7.49\,dB lower than ours. 

\vspace{-14pt}
\paragraph{Forgetting.}
Seq.~ChronoGS is the only method that deteriorates between arrival and final recall.  As every subsequent update modifies the model and causes it to forget about the previous visits, its PSNR decreases by 4.16 to 8.67\,dB at final recall, ultimately falling 4.56 to 6.90\,dB below ours. Fig.~\ref{fig:wat} illustrates this behaviour of Seq.~ChronoGS on Can, Tissue and Tomato scenes of our Multi-State dataset. Meanwhile, our method can recover the selected earlier visits after the final update.

\vspace{-14pt}
\paragraph{Geometric Error.} Image metrics alone do not reveal whether the reconstructed geometry is accurate, so we evaluate the rendered depth of foreground objects on the synthetic scenes of Multi-State. This metric evaluates rendered depth rather than complete surfaces or mesh accuracy. We report results in Table~\ref{tab:hard-geometry} and provide implementation details in the supplementary material. 
At arrival, our method achieves a mean depth error of 44.59\,mm, compared with 115.32\,mm for Seq.~LTGS and 144.19\,mm for Seq.~ChronoGS, and the lowest error on three of the four scenes, with Seq.~LTGS performs better on the Plant scene. Although our method and Seq.~ChronoGS differ by only 0.64\,dB in arrival PSNR on these scenes, as shown in Table~\ref{tab:overview}, our depth error is 3.2$\times$ lower. Moreover, as Seq.~ChronoGS struggles with forgetting, its error increases to 411.34\,mm at final recall. 

\subsection{Analysis and Ablations}
\label{sec:views}\label{sec:ablations}
\paragraph{Change localisation.}
Inaccurate change support can either prevent necessary updates or alter persistent surfaces. We therefore examine how each step in obtaining this support affects reconstruction quality. ReScene obtains change support in two steps, where it localises changes as pixel masks and then attributes these masks to Gaussians through compositing weights. Using Multi-State, we replace each step in turn while keeping the rest of the update pipeline unchanged. As shown in Table~\ref{tab:localisation}, replacing our photometric masks with the DINOv2-based masks of CL-Splats~\cite{clsplats} or the SAM-based masks of LTGS~\cite{ltgs} reduces PSNR by 0.55 and 2.01 dB, respectively on the real scenes. These results indicate that our photometric masks localise changes more effectively for reconstruction than the feature-based alternatives. Moreover, applying the projected-centre voting of CL-Splats to its own masks further reduces PSNR by 2.70 dB. Voting ignores the footprint, opacity and occlusion of each Gaussian, whereas compositing weights transfer evidence in proportion to each Gaussian’s contribution to a pixel. The synthetic scenes of Multi-State also show the same trend.

\begin{table}[t]
\centering\footnotesize
\setlength{\tabcolsep}{3pt}
\caption{\textbf{Change localisation and attribution on the Multi-State dataset}. We compare our proposed mechanism with alternatives derived in CL-Splat~\cite{clsplats} and LTGS~\cite{ltgs}.}
\vspace{-5pt}
\label{tab:localisation}
\resizebox{\columnwidth}{!}{%
\begin{tabular}{@{}llcccc@{}}
\toprule
& & \multicolumn{2}{c}{Real} & \multicolumn{2}{c}{Synthetic}\\
\cmidrule(lr){3-4}\cmidrule(l){5-6}
Change masks & Attribution
& PSNR$\uparrow$ & SSIM$\uparrow$
& PSNR$\uparrow$ & SSIM$\uparrow$\\
\midrule
LTGS      & Compositing & 23.33 & 0.855 & 30.60 & 0.947\\
CL-Splats & Voting      & 22.09 & 0.840 & 31.25 & 0.953\\
CL-Splats & Compositing & 24.79 & 0.870 & 31.63 & 0.954\\
\midrule
Ours      & Compositing & \textbf{25.34} & \textbf{0.872}
          & \textbf{32.50} & \textbf{0.957}\\
\bottomrule
\end{tabular}}
\end{table}
\begin{figure}[t]
\centering
\includegraphics[width=\columnwidth]{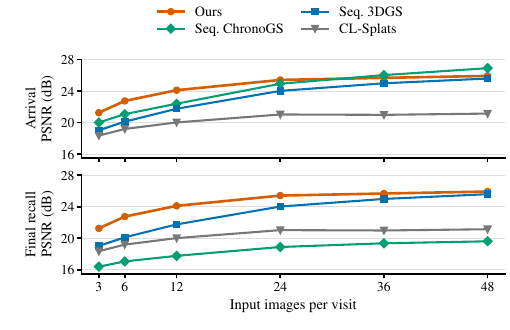}
\vspace{-10pt}
\caption{\textbf{View-count sensitivity on real Multi-State.} Arrival and final-recall PSNR across four scenes, including initialisation. Differences in the text use unrounded means. The horizontal axis gives the number of current images at each revisit.}
\label{fig:views}
\vspace{-10pt}
\end{figure}

\vspace{-12pt}
\paragraph{Number of Images per Visit.} We evaluate subsets of 3 to 48 images per revisit on the real scenes of Multi-State, where each subset extends the preceding one with additional images. The results are shown in Fig.~\ref{fig:views}. Our method achieves the highest arrival PSNR with 3 to 24 images and the highest final-recall PSNR at every evaluated view count. The averaged PSNR generally improves with more images for all methods. With 36 and 48 images, Seq.~ChronoGS achieves higher arrival PSNR, but its final-recall PSNR remains 6.29\,dB below ours at both view counts. At 48 images, Seq.~3DGS approaches our final-recall quality, trailing by 0.33\,dB, but requires 2.63$\times$ the history storage. These results show that our reconstruction-quality advantage is most pronounced under sparse observations, while its storage advantage over checkpointing persists with more views. 


\begin{table}[t]
\centering\footnotesize
\setlength{\tabcolsep}{3pt}
\caption{\textbf{Component ablations on Multi-State.} Each row removes one component from our proposed method. Reported results exclude the initial visit, and are averaged across all scenes.}
\label{tab:component-ablation}
\vspace{-5pt}
\begin{tabular*}{\columnwidth}{@{\extracolsep{\fill}}lcccccc@{}}\toprule
 & \multicolumn{3}{c}{Synthetic} & \multicolumn{3}{c}{Real}\\
\cmidrule(lr){2-4}\cmidrule(l){5-7}
Removed & PSNR$\uparrow$ & SSIM$\uparrow$ & MiB$\downarrow$ & PSNR$\uparrow$ & SSIM$\uparrow$ & MiB$\downarrow$\\\midrule
Births      & 27.82 & 0.929 & 271.5 & 23.44 & 0.851 & 120.2\\
Suppression & 28.64 & 0.933 & 341.8 & 22.76 & 0.846 & 155.7\\
Refinement  & 30.67 & 0.952 & 186.5 & 22.14 & 0.837 & 126.8\\
Splitting   & 31.49 & 0.950 & 308.9 & 24.99 & 0.866 & 141.4\\\midrule
None (full) & 32.11 & 0.955 & 352.0 & 25.16 & 0.870 & 163.6\\
\bottomrule
\end{tabular*}
\vspace{-10pt}
\end{table}
\vspace{-11pt}
\paragraph{Ablation Study.} We remove each proposed component inturn: Gaussian birth, suppression, refinement and splitting. For each removal, we train and evaluate the reconstructed scene using the Multi-State dataset, and the results are shown in Table~\ref{tab:component-ablation}. We exclude the initial visit from evaluation, which these components do not affect. Each removal generally results in performance degradation. For example, removing births reduces PSNR by 4.29 and 1.72\,dB on the synthetic and real scenes, respectively, as surfaces that appear between visits can no longer be represented. Removing suppression causes Gaussians associated with moved or removed surfaces to continue occluding the new content. Refinement contributes most on the real scenes, as real scenes can differ in lighting across visits. Removing refinement also reduces total history storage in both scenes, reflecting the storage--quality trade-off of the resulting update trajectories. Splitting provides smaller gains of 0.62 and 0.18\,dB. 

\vspace{-12pt}
\paragraph{Limitations.}
When the evidence guiding an update in \method{} is unreliable, ghosting artefacts may appear in rendered images. 
Occlusion and appearance variation can corrupt the change support and misguide suppression, retaining outdated Gaussians alongside new content. Depth errors can misplace births, causing them to duplicate existing surfaces. 
\vspace{-4pt}
\section{Conclusion}
\vspace{-4pt}
We introduced ReScene for scene chronology from sparse revisits. It attributes image disagreement to individual Gaussians to guide selective, immutable updates, so every visit is recovered exactly without past images. Our Multi-State benchmark covers non-rigid deformation and partial material removal with ground-truth geometry. ReScene achieves quality comparable to 3DGS checkpointing with 2.5--4.8$\times$ less storage, and 3.2$\times$ lower depth error than Seq.\ ChronoGS at comparable PSNR.

\clearpage
\onecolumn
\pdfbookmark[0]{Supplementary Material}{supplement-start}
\begin{center}
{\Large\bfseries\thetitle\par}
\vspace{0.5em}
{\large Supplementary Material\par}
\end{center}
\raggedbottom
\setcounter{section}{0}
\setcounter{subsection}{0}
\setcounter{table}{0}
\setcounter{figure}{0}
\setcounter{equation}{0}
\renewcommand{\thesection}{S\arabic{section}}
\renewcommand{\thetable}{S\arabic{table}}
\renewcommand{\thefigure}{S\arabic{figure}}
\renewcommand{\theequation}{S\arabic{equation}}
\renewcommand{\theHsection}{supp.\arabic{section}}
\renewcommand{\theHtable}{supp.\arabic{table}}
\renewcommand{\theHfigure}{supp.\arabic{figure}}
\renewcommand{\theHequation}{supp.\arabic{equation}}
\setcounter{dbltopnumber}{4}
\renewcommand{\dbltopfraction}{0.85}
\setcounter{tocdepth}{1}
\addtocontents{toc}{\protect\setcounter{tocdepth}{1}}

This supplementary material provides the method details, benchmark construction, experimental settings and additional results that support the main paper. 

\begingroup
\small
\tableofcontents
\endgroup

Section~\ref{sec:supp-method-details} details the components of ReScene, namely evidence accumulation, birth placement and splitting together with a study of the birth limit, the selection of Gaussians for refinement, and the optimisation settings. Section~\ref{sec:supp-data} describes the construction of the synthetic and real parts of Multi-State. Section~\ref{sec:supp-impl-details} specifies the evaluated scenes and the initialisation and update protocol shared by all methods, and Section~\ref{sec:supp-baselines} describes how each baseline is adapted to sequential revisits. The remaining sections support the experiments of the main paper. Section~\ref{sec:supp-geometry} defines the foreground depth metric reported in Table~2 of the main manuscript. Section~\ref{sec:supp-components} reports the component ablations for each scene, and Section~\ref{sec:supp-view-budget} provides the numerical results and storage underlying the view-count study in Fig.~4 of the main manuscript. Finally, Section~\ref{sec:supp-qualitative} presents additional qualitative comparisons on all datasets, together with a failure case on the Plant scene.



\twocolumn
\iftoggle{cvprfinal}{}{\linenumbers}

\section{Methodology Details}
\label{sec:supp-method-details}
\subsection{Evidence accumulation.}
The change and stable masks identify pixels that disagree or agree with the preceding reconstruction. To attribute this evidence to individual Gaussians, we sum each Gaussian's compositing contributions within the two masks, as defined in Sec.3.3 of the main paper.

We compute these sums using the backward pass of the renderer. With geometry and opacity fixed, we assign an auxiliary scalar attribute $c_i$ to each Gaussian and render it using the usual compositing weights. For a mask $M$, the derivative of the masked rendering is
\[
\frac{\partial}{\partial c_i}
\sum_p M(p)\sum_j a_j(p)c_j
=
\sum_p M(p)a_i(p),
\]
where $p$ indexes pixels and $a_i(p)$ is the compositing weight of Gaussian $i$, including opacity and transmittance. The derivative therefore accumulates the Gaussian's contribution within the mask. Applying this operation to the change and stable masks and averaging across the current views yields $C_i^t$ and $S_i^t$, respectively. These gradients only measure contributions and do not update the Gaussian attributes. The resulting evidence remains fixed during optimisation of the visit.

\subsection{Birth placement, splitting and capacity}
\label{sec:supp-births}
Section 3.4.2 of the main paper introduces new Gaussians, which we called \textit{births}, to represent surfaces missing from the preceding reconstruction. Their placement uses depth estimates from the current visit, while subsequent splitting provides additional detail where needed. Here, we specify the placement thresholds and splitting settings, then examine how the maximum number of births per visit affects reconstruction quality and history storage.

\paragraph{Placement thresholds.}
We initialise births by back-projecting changed pixels using the estimated depth of the current visit. A candidate must have MASt3R confidence above 1.5 and depth support from at least two current views, including its source view. A supporting view must contain the reprojected point within its confident change mask, and the point's camera-space depth must agree with the estimated depth within 3\%. The candidate depth must also differ from the preceding rendered depth by more than 3\% of the estimated depth. Differences in either direction are accepted, allowing births both in front of existing geometry and behind removed surfaces.

We sample candidates at a pixel stride of two and reject non-finite positions. To exclude distant outliers, candidates must lie within three times the preceding model's 95th-percentile radial extent, measured from its mean centre. We randomly select at most 30,000 candidates using the fixed experiment seed. Their initial colours are sampled from the source images, and their isotropic scales are derived from depth and focal length. Each birth starts with opacity 0.1 and identity rotation.

\paragraph{Splitting settings.}
We split births every 500 steps from step 1,000 to step 4,500, giving eight splitting events during each 5,000-step update. Between events, we accumulate the gradient norm of each birth's position-offset parameter. Births with opacity above 0.05 and positive accumulated gradient are ranked by this value. There is no additional fixed gradient threshold. Each event replaces up to 10\% of the current births with two children per parent, subject to the limit of 60,000 active births. Existing Gaussians are never split.

The children are displaced in opposite directions along the parent's longest principal axis by half its scale along that axis. Their scales are divided by 1.6, and each receives opacity $1-\sqrt{1-\alpha_{\mathrm{parent}}}$. Optimiser moments are retained for unsplit births and initialised to zero for the children. The accumulated gradients are reset after each event. Births with opacity below 0.01 are discarded when saving the update.

\begin{table*}[t]
\centering\footnotesize
\setlength{\tabcolsep}{4pt}
\caption{\textbf{Birth limits on Multi-State.}
Each limit bounds the total number of active births per visit, including children created by splitting. Quality is averaged over later visits, excluding initialisation, and equally across scenes. On the synthetic scenes, PSNR$_\text{FG}$ and PSNR$_\text{BG}$ denote foreground and background PSNR computed using ground-truth object masks. All other quality metrics are computed on whole images. Storage is the mean uncompressed history size per scene in MiB. The shaded row is the default limit. Limits above 64,306 never become active.}
\label{tab:birth-limit}
\begin{tabular}{@{}r cccccc c cccc@{}}
\toprule
\multirow{2}{*}{Limit} & \multicolumn{6}{c}{Synthetic} && \multicolumn{4}{c}{Real}\\
\cmidrule(lr){2-7}\cmidrule(l){9-12}
& PSNR$\uparrow$ & PSNR$_\text{FG}\uparrow$ & PSNR$_\text{BG}\uparrow$
& SSIM$\uparrow$ & LPIPS$\downarrow$ & Storage$\downarrow$
&& PSNR$\uparrow$ & SSIM$\uparrow$ & LPIPS$\downarrow$ & Storage$\downarrow$\\
\midrule
5,000  & 31.16 & 23.00 & 35.14 & 0.948 & 0.081 & 286.9 && 24.50 & 0.859 & 0.359 & 124.1\\
10,000 & 31.54 & 23.63 & 35.23 & 0.950 & 0.077 & 302.2 && 24.76 & 0.863 & 0.352 & 128.0\\
20,000 & 31.91 & 24.27 & 35.29 & 0.953 & 0.073 & 323.4 && 24.91 & 0.865 & 0.345 & 136.3\\
30,000 & 32.06 & 24.54 & 35.31 & 0.954 & 0.072 & 339.4 && 25.02 & 0.867 & 0.340 & 144.6\\
40,000 & 32.11 & 24.65 & 35.31 & 0.955 & 0.071 & 346.4 && 25.15 & 0.869 & 0.335 & 152.4\\
50,000 & 32.11 & 24.67 & 35.29 & 0.955 & 0.071 & 350.0 && 25.16 & 0.870 & 0.334 & 158.8\\
\rowcolor{gray!15}
60,000 & 32.12 & 24.66 & 35.30 & 0.955 & 0.071 & 351.9 && 25.16 & 0.870 & 0.333 & 163.6\\
70,000 & 32.12 & 24.69 & 35.30 & 0.955 & 0.070 & 352.4 && 25.16 & 0.870 & 0.333 & 165.7\\
80,000 & 32.10 & 24.67 & 35.28 & 0.955 & 0.071 & 352.3 && 25.15 & 0.870 & 0.333 & 165.8\\
90,000 & 32.12 & 24.68 & 35.30 & 0.955 & 0.070 & 352.2 && 25.16 & 0.870 & 0.333 & 165.7\\
\bottomrule
\end{tabular}
\end{table*}
\paragraph{Birth-limit analysis.}
The birth limit bounds the number of active births at each visit, including children created by splitting. Each update first places at most 30,000 births, or fewer if the limit is lower. Births are then split every 500 steps from step 1,000 to step 4,500 of the 5,000-step update, giving eight splitting events. Starting after 1,000 steps allows births to accumulate meaningful positional gradients, and stopping at step 4,500 leaves the final steps for the newest children to converge. At each event, up to 10\% of the current births are each replaced by two children, so the count increases by at most 10\%. Starting from 30,000, the count can therefore reach at most 64,306 births after the eight events. A limit of 60,000 thus truncates only the final splitting event, whereas limits of 70,000 or more are never reached. In practice, fewer births may be created, as only births with opacity above 0.05 and positive accumulated gradient are split, and births with opacity below 0.01 are discarded when saving.

We vary the limit from 5,000 to 90,000 on both subsets of Multi-State, running complete histories with all other settings fixed, including the initial placement cap, splitting schedule, training views, optimisation budget and seed. Each variant updates its own preceding reconstruction, so differences propagate through subsequent visits. Table~\ref{tab:birth-limit} reports quality averaged over later visits together with complete-history storage.

Increasing the limit improves reconstruction quality, with diminishing gains beyond 30,000--40,000 births. On the synthetic scenes, raising the limit from 5,000 to 30,000 improves foreground PSNR by 1.54\,dB, whereas background PSNR changes by only 0.17\,dB. As births are placed by back-projecting changed pixels, and existing Gaussians are never split, the additional capacity improves the changed content while leaving persistent content largely unaffected. As ground-truth object masks are unavailable for the real scenes, we report whole-image metrics there, where raising the limit from 5,000 to 40,000 improves PSNR by 0.66\,dB.

Beyond 40,000 births, whole-image PSNR changes by at most 0.02\,dB on both subsets. This is comparable to the differences among the limits of 70,000--90,000, which are never reached and therefore reflect variation between runs rather than an effect of the limit. From 40,000 to 60,000, storage increases by 5.5 and 11.2\,MiB per scene on the synthetic and real scenes, respectively. We use a limit of 60,000 in all other experiments, as it allows the splitting schedule to proceed almost fully and acts as a bound on per-visit storage rather than a constraint on quality. A limit of 40,000 would yield comparable quality with slightly less storage.

\subsection{Selection of Gaussians for refinement.}
Section 3.4.3 of the main paper introduces the refinement set $\mathcal F^t$, which combines Gaussians selected through image-based evidence with a conservative depth-supported subset. For Gaussian $i$ at visit $t$, $C_i^t$ and $S_i^t$ denote its accumulated compositing contributions within the change and stable masks, respectively, averaged across the current views. Please refer to Equation 2 in the main manuscript for the formulation. The image-based selection includes Gaussians for which $C_i^t/(C_i^t+1)$ or $S_i^t/(S_i^t+1)$ exceed $10^{-6}$. This identifies Gaussians that contribute to changed or stable pixels.

For the depth-supported subset, a Gaussian must have opacity above 0.005. In each supporting view, its projected centre must lie within the image, and its camera-space depth must agree with the preceding rendered depth within 3\%. At least two current views must also provide confident depth agreement at that location. Specifically, MASt3R confidence must exceed 1.5, the estimated depth must agree with the preceding rendered depth within 3\%, and the pixel must lie outside the change mask used during depth-proposal generation. A visible observation inside this change mask excludes the Gaussian from the depth-supported subset. We also exclude the suppression candidates identified during that proposal stage. These exclusions apply only to the depth-supported subset. The final refinement set is its union with the image-based selection, so refinement and suppression may still affect the same Gaussian.

\begin{figure*}[t]
\centering\includegraphics[width=\textwidth]{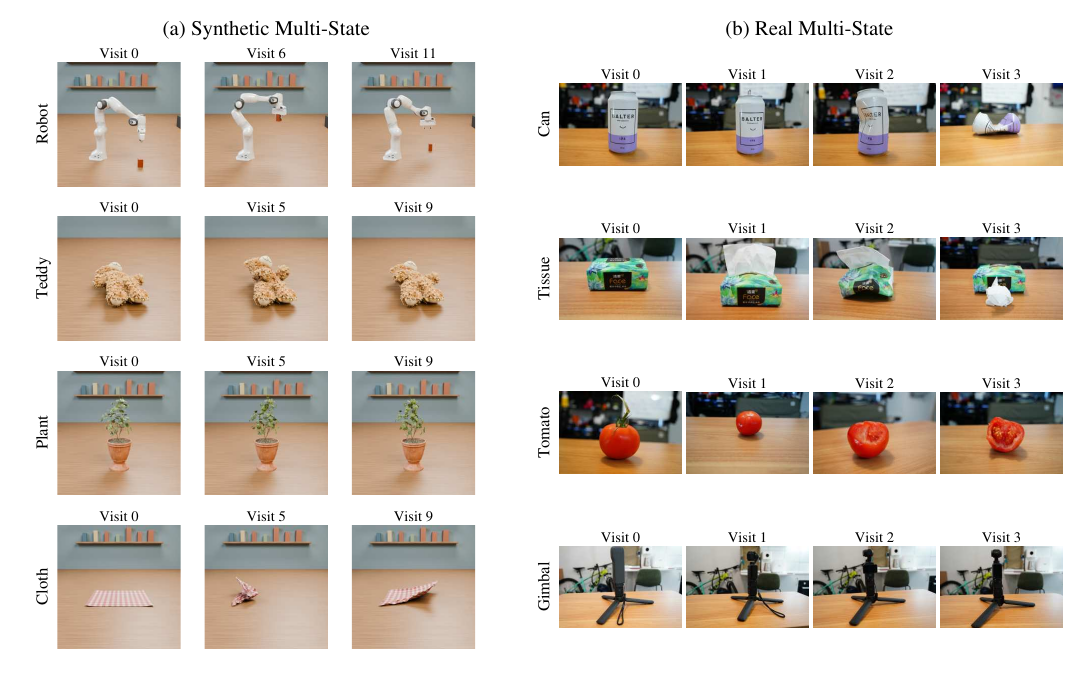}
\caption{Changing states in Multi-State. (a) First, intermediate and final states of Robot, Teddy, Plant and Cloth, rendered from a fixed held-out camera. (b) Can, Tissue, Tomato and Gimbal at visits 0--3, shown using captured held-out images. The real viewpoints are approximately matched across visits, rather than identical.}
\vspace{-10pt}
\label{fig:multistate-states}
\end{figure*}

\subsection{Optimisation settings}
\label{sec:supp-optimisation}
This subsection provides the settings for the update optimisation described in Section 3.5 of the main paper. At each visit, we initialise a fresh Adam optimiser with learning rate $10^{-3}$ and optimise for 5,000 steps. Each step samples one image uniformly from the selected training views of the current visit. Images from previous visits are not used during the update of the current visit.

We jointly optimise the visit-specific MLP, the selected Gaussian offsets, the birth parameters and the appearance correction. The MLP is initialised independently at each visit, with its output layer set to zero. Refinement offsets and the appearance log-gain and bias are also initialised to zero. Existing Gaussian rotations and higher-order colour coefficients kept frozen, while birth higher-order colour coefficients were set to zero and remain frozen. We clip the total gradient norm to 10 before each optimiser step to ensure stablility, and the parameters stored for earlier visits remain unchanged.

The photometric loss $\mathcal{L}_\text{photo}$ combines absolute RGB error and SSIM dissimilarity with weights 0.8 and 0.2, respectively. Each masked loss is normalised by its mask weight. The change, stable and refinement photometric terms and the image-preservation term each have unit weight. Change evidence, supervision masks and the refinement selection are computed before optimisation and remain fixed throughout the update. Birth splitting follows the schedule described in Sec.3.4.2 in the main manuscript.

\section{The Multi-State benchmark}
\label{sec:supp-data}
Multi-State comprises synthetic scenes, which provide ground-truth geometry at every state, and real scenes, which contain the imperfections of real-world capture. Each state is a static configuration of a scene, and consecutive states differ by rigid or non-rigid changes.

\subsection{Synthetic scenes}
The synthetic part contains four sequences, and Figure~\ref{fig:multistate-states}(a) shows the first, intermediate and final states of each scene. Robot has 12 states and shows arm and gripper articulation while transporting a payload. Teddy, Plant and Cloth have 10 states each. Teddy bends, twists and compresses, Plant moves its branches and leaves around a fixed pot, and Cloth is lifted, twisted, folded and reopened with self-contact. Each state is a manually specified configuration, so the sequences represent neither measured time intervals nor calibrated material behaviour.

The Robot sequence uses the Franka Panda model from MuJoCo Menagerie (Apache-2.0), and the Teddy sequence uses a model from Google Scanned Objects, obtained through Kubric (CC~BY-SA~4.0). The Plant sequence uses the Potted Plant~01 model, and the Cloth sequence uses the Fabric Pattern~07 texture, both from Poly Haven (CC0). We created the cloth mesh, payload, background and all motions. Leaf transparency is converted to explicit geometry so that image silhouettes agree with the geometric annotations. Cameras, lighting, exposure and background remain fixed within each sequence.

For all scenes, each state is rendered from 60 calibrated cameras at $800\times800$ pixels, together with ground-truth depth, normals, meshes and foreground masks. Every eighth view, starting from view 0, is held out for testing, leaving 52 views per state for training. The synthetic part therefore contains 42 states, 2,520 images and 336 held-out views.

\subsection{Real scenes}
\label{sec:real-multistate}
\paragraph{Capture.}
The real subset contains four objects, Can, Tissue, Tomato and Gimbal, each captured in four physical states, as shown in Figure~\ref{fig:multistate-states}(b). The can is opened and progressively crushed, a tissue is pulled out of its packet and crumpled, material is removed from the tomato, and the gimbal and its cover are changed. Images were captured with a Sony $\alpha$7~III at $6000\times4000$ pixels and focal lengths of 28--29\,mm. Each state is captured as a separate static scene from viewpoints that vary in direction and elevation, and the support surface and room background are retained. Similar to the synthetic subset, we take every eighth registered image of each state as held-out view. As summarised in Table~\ref{tab:real-capture}, the real subset consists of 16 states, 1,658 images and 214 held-out views.
\vspace{-10pt}
\paragraph{Per-state camera estimation.}
Each state is reconstructed independently with COLMAP~\cite{colmap}, using exhaustive matching with geometric verification and incremental mapping with bundle adjustment. SIFT features are extracted at a maximum image dimension of 3,200 pixels, with up to 16,384 features per image. Images of a state that share a focal length share an \texttt{OPENCV} camera model, whose focal length and distortion are refined.
\vspace{-10pt}
\paragraph{Cross-visit registration.}
As each state is reconstructed independently, its coordinate frame has an arbitrary origin, orientation and scale. We therefore align the states of each scene with a similarity transform estimated from SIFT correspondences on unchanged background structure.


\begin{table}[t]\centering\footnotesize
\setlength{\tabcolsep}{3pt}
\caption{Real capture collection and evaluation subset. }
\label{tab:real-capture}
\begin{tabular}{lcccc}\toprule
Scene & States & Train Views & Test Views\\\midrule
Can & 4  & 337 & 45 \\
Tissue & 4  & 461 & 59 \\
Tomato & 4  & 462 & 59 \\
Gimbal & 4  & 398 & 51 \\
\bottomrule\end{tabular}\end{table}

\section{Implementation Details}
\label{sec:supp-impl-details}

\subsection{Evaluated scenes}
We evaluate all four synthetic sequences and real sequences of Multi-State, which is described in previous section. We further evaluate ten scenes of WAT~\cite{clnerf}, with 51 states, 41 updates and 377 held-out images. The community scene is excluded as the sequential LTGS adaptation could not be validated on it in our experiments. On ChronoScene~\cite{chronogs}, we evaluate twelve scenes with 42 states, 30 updates and 999 held-out images, excluding three training images in Canteen and Lawncourt that lack a camera registration. For WAT and ChronoScene, we use the camera registrations and held-out images provided by the original benchmark, and all methods are evaluated on the same scenes of every dataset.

\subsection{3DGS Initialisation and Updates}

Across all datasets, only the \textit{initial} reconstruction of each scene is trained on all training images of the first visit. On synthetic Multi-State, it is initialised from points back-projected from the ground-truth depth of these views, and the ground-truth geometry of later states is used only for evaluation as mentioned in the main manuscript. On the real (Multi-State and WAT) scenes and ChronoScene, it is initialised from SfM points restricted to tracks observed in at least two initial training images. Methods including ReScene, sequential 3DGS, sequential LTGS and CL-Splats share this initial model, which is fitted for 30,000 iterations. Meanwhile, sequential ChronoGS uses 40,000 iterations for initial fit, following its original protocol.

Across all datasets, each later state provides at most 24 training images. These images were selected by farthest-point sampling of camera centres. Starting from the first image in lexicographic order, or on WAT from the three evenly spaced frames of the original sparse subset, we repeatedly add the camera farthest from those already selected. The selection depends only on camera positions, and the selected images and held-out cameras are identical across methods. Each update runs for 5,000 steps of optimisation.

\section{Baseline optimisation and sequential adaptations}
\label{sec:supp-baselines}

This section provides the implementation details for the baselines introduced in Sec.4.1.2 of the main paper. We evaluate all methods using the same registered cameras, selected training images and held-out views. Each revisit supplies up to 24 current training images and a budget of 5,000 optimisation steps. The methods differ in their initialisation, use of synthetic replay and representation of earlier visits. These adaptations therefore match the observation setting and update budget, rather than computational cost.

\subsection{Sequential 3DGS}

\paragraph{Sequential adaptation.}
We extend static 3DGS to sequential visits by initialising each update from the preceding fitted model. All Gaussian positions, scales, rotations, opacities and spherical-harmonic colour coefficients are optimised using the current visit's images. We do not provide earlier captured images, synthetic replay or MASt3R-based birth proposals. This baseline tests whether conventional Gaussian fine-tuning can accommodate successive scene changes.
\vspace{-10pt}
\paragraph{Optimisation.}
Each visit uses a fresh Adam optimiser with the standard 3DGS parameter-group learning rates and nominal 30,000-step learning-rate schedule. We execute the first 5,000 steps of this schedule for each update. The reconstruction loss combines $\mathrm{L}_1$ error and SSIM dissimilarity with weights 0.8 and 0.2, respectively. Native densification and pruning occur every 100 steps from step 600 through step 4,900. Opacity resets and world-size pruning are disabled.
\vspace{-10pt}
\paragraph{Historical retrieval.}
Fine-tuning modifies the preceding reconstruction and does not preserve its earlier values. We therefore save a complete renderable model after each visit. Final recall loads the corresponding saved model, and the reported history storage includes the initial model and all subsequent checkpoints.

\subsection{Sequential ChronoGS}

\paragraph{Sequential adaptation.}
ChronoGS~\cite{chronogs} represents multiple visits within a shared temporal model. We adapt its joint fitting procedure by introducing visits sequentially. Each update starts from the preceding model and appends 16-dimensional anchor features and a global embedding for the new visit. These entries are initialised from those of the preceding visit.

We then optimise the shared parameters and Gaussian topology using only the current visit's images. Earlier temporal features remain trainable. We introduce no replay, compact adapters or additional persistence penalty. This retains the native shared representation while restricting captured-image access to the current visit.
\vspace{-10pt}
\paragraph{Optimisation.}
The initial reconstruction is fitted using the native 40,000-step ChronoGS procedure. Each subsequent visit uses a fresh Adam optimiser and restarts the native 40,000-step learning-rate schedule, executing 5,000 update steps. The native losses and parameter-group settings are retained. This initialisation differs from the standard 3DGS initial model shared by ReScene, sequential 3DGS and sequential LTGS.
\vspace{-10pt}
\paragraph{Historical retrieval.}
Final recall queries the earlier visit indices in the final shared model. Later updates can therefore affect earlier reconstructions through changes to shared parameters and topology. We count the final shared model in storage, rather than a collection of per-visit checkpoints.

\subsection{Sequential LTGS}

\paragraph{Sequential adaptation.}
LTGS~\cite{ltgs} uses object templates to reconstruct scene changes from sparse observations. We adapt its multi-visit pipeline so that each update begins from its own preceding reconstruction and processes only the newly available visit. We retain its change detection, instance matching, geometry initialisation and local refinement stages. Camera localisation is skipped because registered cameras are supplied consistently to all methods.

The current visit supplies captured images for the update. To retain previously reconstructed content, the preceding model generates synthetic replay images at the initial and earlier visits' training poses. Earlier camera poses are therefore retained, but earlier captured RGB images are not used. Future visits are unavailable throughout the update.
\vspace{-10pt}
\paragraph{Optimisation.}
Sequential LTGS starts from the same standard 3DGS initial model as ReScene and sequential 3DGS. At each revisit, we run the released refinement procedure with a budget of 5,000 optimisation steps, retaining its native optimiser settings and losses. Synthetic replay remains enabled through the previous-viewpoint option. Replay and the preceding matching stages incur additional computation, so the common step budget does not imply equal runtime. Earlier WAT runs perform 4,999 optimiser updates because of the original terminal-step convention. Subsequent runs perform 5,000 updates.

We apply integration repairs for numeric mask labels, visibility handling and variable camera dimensions. For images with different dimensions, masks are mapped to the corresponding image grid using nearest-label sampling before track lookup. These repairs preserve label identities and matching thresholds.
\vspace{-10pt}
\paragraph{Historical retrieval.}
Each fitted visit is exported as a complete renderable model and retained for final recall. This separates historical reconstruction from subsequent changes to the working model. The reported storage includes the initial model and every saved visit. Thus, sequential LTGS uses synthetic replay during updating and checkpoint retrieval during historical evaluation.

\section{Geometric accuracy evaluation}
\label{sec:supp-geometry}

This section describes the foreground depth evaluation reported in Sec. 4.2.3 and Table 2 of the main paper. We compare reconstructed depth with ground-truth depth to assess geometric accuracy beyond rendered appearance. Evaluation uses held-out cameras from Robot, Teddy, Plant and Cloth in synthetic Multi-State. Each reconstruction is evaluated immediately after its visit is fitted.
\vspace{-10pt}
\paragraph{Rendered depth.}
We compute depth using the compositing weights of the Gaussian renderer. Let $z_i$ denote the depth of Gaussian $i$'s centre along the camera's optical axis, and let $a_i(p)$ denote its compositing weight at pixel $p$, including opacity and transmittance. Accumulated opacity $A(p)$ and rendered expected depth $\hat D(p)$ are
\begin{equation}
 A(p)=\sum_i a_i(p),
 \qquad
 \hat D(p)=\frac{\sum_i a_i(p)z_i}{A(p)}.
\end{equation}
The resulting depth is the opacity-normalised expected camera-space depth of the contributing Gaussian centres. It is not the distance along the viewing ray or the depth of the first surface intersection.

\paragraph{Foreground support.}
We use ground-truth object masks to exclude the background. Let $F$ contain foreground pixels with finite, positive ground-truth depth $D_{\mathrm{GT}}(p)$. We retain pixels whose reconstructed depth is finite and positive and whose accumulated opacity is at least 0.5,
\begin{equation}
 Q=\left\{
 p\in F \;\middle|\;
 A(p)\geq0.5,\;
 0<\hat D(p)<\infty
 \right\}.
\end{equation}
Although the foreground mask is shared across methods, the opacity condition can retain different pixels. We therefore also measure foreground coverage to indicate how much of the object contributes to the error.

\begin{table}[t]
\centering\footnotesize
\setlength{\tabcolsep}{2pt}
\caption{\textbf{Scene-wise component ablations on synthetic Multi-State.}
PSNR and SSIM exclude initialisation. Storage is the complete
uncompressed history in MiB. Bold marks the best quality per scene.}
\label{tab:component-synthetic}
\begin{tabular*}{\columnwidth}{@{\extracolsep{\fill}}llccc@{}}
\toprule
Scene & Variant & PSNR$\uparrow$ & SSIM$\uparrow$ & MiB$\downarrow$\\
\midrule
Robot & Full & \textbf{35.71} & \textbf{0.9746} & 386.3\\
 & No births & 26.35 & 0.9420 & 285.6\\
 & No suppression & 28.54 & 0.9471 & 382.4\\
 & No refinement & 32.65 & 0.9699 & 204.2\\
 & No splitting & 34.95 & 0.9729 & 334.7\\
\midrule
Teddy & Full & \textbf{31.84} & \textbf{0.9501} & 306.5\\
 & No births & 29.03 & 0.9240 & 224.4\\
 & No suppression & 29.00 & 0.9239 & 291.9\\
 & No refinement & 29.75 & 0.9363 & 158.3\\
 & No splitting & 31.12 & 0.9415 & 264.8\\
\midrule
Plant & Full & 30.07 & 0.9397 & 306.9\\
 & No births & 29.71 & 0.9286 & 276.1\\
 & No suppression & 29.48 & 0.9402 & 275.1\\
 & No refinement & \textbf{31.04} & \textbf{0.9554} & 152.6\\
 & No splitting & 29.88 & 0.9376 & 291.7\\
\midrule
Cloth & Full & \textbf{30.83} & \textbf{0.9553} & 408.2\\
 & No births & 26.20 & 0.9205 & 300.0\\
 & No suppression & 27.55 & 0.9188 & 417.9\\
 & No refinement & 29.22 & 0.9450 & 230.8\\
 & No splitting & 30.03 & 0.9476 & 344.2\\
\bottomrule
\end{tabular*}
\end{table}

\paragraph{Error computation and aggregation.}
For each held-out view, we compute mean absolute depth error and foreground coverage as
\begin{equation}
 \mathrm{MAE}
 =
 \frac{1}{|Q|}
 \sum_{p\in Q}
 \left|\hat D(p)-D_{\mathrm{GT}}(p)\right|,
 \qquad
 \mathrm{Coverage}=\frac{|Q|}{|F|}.
\end{equation}
We convert errors from metres to millimetres for reporting. Lower MAE indicates more accurate foreground depth. We average view errors within each visit, then average visits within each scene, including initialisation. The overall mean weights the four scenes equally. Mean foreground coverage exceeds 99.99\% for every evaluated method.

\begin{table}[t]
\centering\footnotesize
\setlength{\tabcolsep}{2pt}
\caption{\textbf{Scene-wise component ablations on real Multi-State.}
PSNR and SSIM exclude initialisation. Storage is the complete
uncompressed history in MiB. Bold marks the best quality per scene.}
\label{tab:component-real}
\begin{tabular*}{\columnwidth}{@{\extracolsep{\fill}}llccc@{}}
\toprule
Scene & Variant & PSNR$\uparrow$ & SSIM$\uparrow$ & MiB$\downarrow$\\
\midrule
Can & Full & \textbf{23.59} & \textbf{0.8633} & 160.7\\
 & No births & 22.20 & 0.8494 & 118.7\\
 & No suppression & 21.84 & 0.8427 & 151.9\\
 & No refinement & 22.06 & 0.8401 & 124.1\\
 & No splitting & 23.40 & 0.8602 & 139.2\\
\midrule
Tissue & Full & \textbf{23.82} & \textbf{0.8548} & 167.9\\
 & No births & 20.84 & 0.8149 & 120.8\\
 & No suppression & 20.98 & 0.8214 & 158.5\\
 & No refinement & 20.39 & 0.8158 & 128.5\\
 & No splitting & 23.46 & 0.8459 & 144.2\\
\midrule
Tomato & Full & \textbf{26.94} & 0.8787 & 142.3\\
 & No births & 26.15 & 0.8739 & 105.2\\
 & No suppression & 25.14 & 0.8637 & 138.8\\
 & No refinement & 23.22 & 0.8444 & 114.5\\
 & No splitting & 26.88 & \textbf{0.8790} & 122.4\\
\midrule
Gimbal & Full & \textbf{26.30} & \textbf{0.8814} & 183.5\\
 & No births & 24.57 & 0.8668 & 136.1\\
 & No suppression & 23.07 & 0.8573 & 173.7\\
 & No refinement & 22.90 & 0.8462 & 139.8\\
 & No splitting & 26.20 & 0.8803 & 159.7\\
\bottomrule
\end{tabular*}
\end{table}
\vspace{-10pt}
\paragraph{Interpretation.}
This metric measures foreground depth accuracy from the held-out viewpoints. It does not assess complete surfaces, unobserved geometry or mesh completeness. It also differs from the harmonic-depth approximation used to prepare birth proposals and refinement evidence. Ground-truth depth from later visits and foreground masks are reserved for evaluation. Only initial training-view depth is used to initialise the reconstruction, as described in the dataset construction section.

\begin{table*}[t]\centering\footnotesize
\caption{Equal-scene quality across four real Multi-State scenes. A and R denote arrival and final recall, respectively. \textit{All} includes initial model while \textit{Later} excludes it. SSIM and LPIPS refer to all visits (including the initial model).}
\label{tab:real-view-metrics}
\begin{tabular}{lrrrrrrrrr}\toprule
Method & Views & All A & All R & Later A & Later R & SSIM A & SSIM R & LPIPS A & LPIPS R\\\midrule
Ours & 3 & 21.26 & 21.26 & 19.64 & 19.64 & 0.8103 & 0.8103 & 0.3824 & 0.3824\\
Ours & 6 & 22.75 & 22.75 & 21.63 & 21.63 & 0.8356 & 0.8356 & 0.3584 & 0.3584\\
Ours & 12 & 24.11 & 24.11 & 23.44 & 23.44 & 0.8574 & 0.8574 & 0.3387 & 0.3387\\
Ours & 24 & 25.40 & 25.40 & 25.17 & 25.17 & 0.8732 & 0.8732 & 0.3221 & 0.3221\\
Ours & 36 & 25.67 & 25.67 & 25.52 & 25.52 & 0.8770 & 0.8770 & 0.3184 & 0.3184\\
Ours & 48 & 25.91 & 25.91 & 25.85 & 25.85 & 0.8795 & 0.8795 & 0.3165 & 0.3165\\
\midrule
Seq. ChronoGS & 3 & 20.02 & 16.41 & 17.91 & 16.46 & 0.7292 & 0.6393 & 0.4378 & 0.5297\\
Seq. ChronoGS & 6 & 21.08 & 17.08 & 19.32 & 17.34 & 0.7776 & 0.6937 & 0.3938 & 0.4865\\
Seq. ChronoGS & 12 & 22.40 & 17.78 & 21.08 & 18.24 & 0.8201 & 0.7345 & 0.3532 & 0.4492\\
Seq. ChronoGS & 24 & 24.92 & 18.90 & 24.45 & 19.72 & 0.8626 & 0.7737 & 0.3119 & 0.4098\\
Seq. ChronoGS & 36 & 26.02 & 19.37 & 25.90 & 20.33 & 0.8762 & 0.7876 & 0.3005 & 0.3978\\
Seq. ChronoGS & 48 & 26.89 & 19.62 & 27.07 & 20.73 & 0.8850 & 0.7932 & 0.2939 & 0.3945\\
\midrule
Seq. 3DGS & 3 & 19.03 & 19.03 & 16.68 & 16.68 & 0.7189 & 0.7189 & 0.4612 & 0.4612\\
Seq. 3DGS & 6 & 20.13 & 20.13 & 18.13 & 18.13 & 0.7701 & 0.7701 & 0.4114 & 0.4114\\
Seq. 3DGS & 12 & 21.76 & 21.76 & 20.31 & 20.31 & 0.8214 & 0.8214 & 0.3626 & 0.3626\\
Seq. 3DGS & 24 & 24.02 & 24.02 & 23.32 & 23.32 & 0.8606 & 0.8606 & 0.3223 & 0.3223\\
Seq. 3DGS & 36 & 24.98 & 24.98 & 24.60 & 24.60 & 0.8729 & 0.8729 & 0.3105 & 0.3105\\
Seq. 3DGS & 48 & 25.58 & 25.58 & 25.41 & 25.41 & 0.8792 & 0.8792 & 0.3038 & 0.3038\\
\midrule
CL-Splats & 3 & 18.37 & 18.37 & 15.78 & 15.78 & 0.7862 & 0.7862 & 0.4212 & 0.4212\\
CL-Splats & 6 & 19.20 & 19.20 & 16.90 & 16.90 & 0.8042 & 0.8042 & 0.4017 & 0.4017\\
CL-Splats & 12 & 20.03 & 20.03 & 17.99 & 17.99 & 0.8172 & 0.8172 & 0.3904 & 0.3904\\
CL-Splats & 24 & 21.03 & 21.03 & 19.34 & 19.34 & 0.8257 & 0.8257 & 0.3782 & 0.3782\\
CL-Splats & 36 & 20.99 & 20.99 & 19.28 & 19.28 & 0.8261 & 0.8261 & 0.3772 & 0.3772\\
CL-Splats & 48 & 21.15 & 21.15 & 19.49 & 19.49 & 0.8288 & 0.8288 & 0.3745 & 0.3745\\
\midrule
\vspace{-20pt}
\end{tabular}\end{table*}

\section{Scene-wise component ablations}
\label{sec:supp-components}

Table~\ref{tab:component-synthetic} and \ref{tab:component-real} provide the scene-wise results underlying the component ablation in Sec. 4.3.2 of the main paper. Each variant is trained over the complete sequence using its own preceding reconstruction. We average held-out PSNR within each visit and then across subsequent visits, excluding initialisation. The initial model, selected training images and optimisation settings are shared across variants within each scene.
\vspace{-10pt}
\paragraph{Findings.}
The contribution of each component varies across scenes. Removing births causes the largest reduction on Robot and Cloth, while removing suppression substantially reduces quality on Robot and Gimbal. Refinement improves PSNR in seven of the eight scenes, with its largest benefit on Tomato. Plant is the exception, where disabling refinement increases PSNR from 30.07 to 31.04\,dB. This shows that refinement is not uniformly beneficial and that the aggregate improvement does not extend to every scene. Splitting provides smaller gains, particularly on the real scenes, where removing it reduces PSNR by 0.06--0.36\,dB. The full method achieves the highest PSNR on all four real scenes and three synthetic scenes. These single-seed comparisons describe observed differences rather than statistically established effects.

\section{Analysis on Number of Images per Visit}
\label{sec:supp-view-budget}
We evaluate ReScene, sequential ChronoGS, sequential 3DGS and the refined CL-Splats adaptation on our real Multi-State dataset. Each method reconstructs a complete history using 3, 6, 12, 24, 36 or 48 current training views. Every revisit has enough training images for all six settings.
\paragraph{Inputs and training.}
We construct nested subsets using farthest-camera-centre selection. Each subset starts with the first filename in lexicographic order and retains all images from the smaller subset. The 24-view subset is identical to that used in the main experiment. Within each method, the initial model and held-out cameras remain fixed. Initial fitting uses all initial training images. Each update uses 5,000 steps, seed zero and its own preceding reconstruction. ReScene recomputes depth and evidence for each subset. All methods retain the optimiser and topology schedules described in Section~\ref{sec:supp-baselines}. Updates do not use captured images from earlier or future visits. The offline camera-registration limitations and different ChronoGS initial fit remain unchanged. Equal iteration counts do not imply equal computational cost.
\begin{table}[t]
\centering\footnotesize
\setlength{\tabcolsep}{5pt}
\caption{\textbf{Storage and update-training cost by view count.}
Each cell reports mean uncompressed history storage per scene (MiB) /
total update-training time across four scenes (hours).
Time excludes initial fitting, preparation, evaluation and failed attempts.}
\label{tab:real-view-cost}
\begin{tabular}{@{}ccccc@{}}
\toprule
Views & Ours & Seq. ChronoGS & Seq. 3DGS & CL-Splats\\
\midrule
3  & 115.6/0.43 & 98.7/0.49  & 344.1/0.27 & 95.0/0.32\\
6  & 135.7/0.44 & 107.2/0.49 & 358.0/0.26 & 90.3/0.33\\
12 & 153.5/0.45 & 112.8/0.49 & 383.9/0.26 & 89.9/0.34\\
24 & 163.6/0.46 & 108.6/0.48 & 410.5/0.27 & 91.2/0.33\\
36 & 168.2/0.46 & 104.9/0.47 & 430.6/0.27 & 92.0/0.32\\
48 & 167.3/0.46 & 100.4/0.46 & 440.4/0.28 & 91.8/0.32\\
\bottomrule
\end{tabular}
\vspace{-10pt}
\end{table}

\vspace{-10pt}
\paragraph{Quality and retention.}
Table~\ref{tab:real-view-metrics} reports the reconstruction results of different methods using the same set of held-out images. 
ReScene has the highest mean arrival PSNR at 3--24 views and the highest final-recall PSNR at every view count. At 36 and 48 views, ChronoGS has higher arrival PSNR. However, its final-recall PSNR is lower than ReScene's by 6.29 and 6.29\,dB, respectively. ReScene's final-recall advantage over sequential 3DGS narrows from 2.22\,dB at three views to 0.33\,dB at 48. These differences use the unrounded scores. Mean quality generally improves with more views, but CL-Splats declines slightly from 24 to 36 views. These single-seed results do not establish statistical significance.
\paragraph{Storage and cost.}
Table~\ref{tab:real-view-cost} reports the uncompressed storage required to retain all reconstructed visits, including the initial model. ReScene stores its updates, MLP weights and auxiliary arrays, whereas sequential 3DGS stores a complete model for each visit. At 48 views, sequential 3DGS requires 2.63$\times$ ReScene's history storage despite similar final-recall quality. Training time is summed across all updates and excludes initial fitting, depth estimation, selection preparation and evaluation.

\section{Additional qualitative results}
\label{sec:supp-qualitative}
\begin{figure*}[p]
\centering
\includegraphics[width=\textwidth,height=.86\textheight,keepaspectratio]{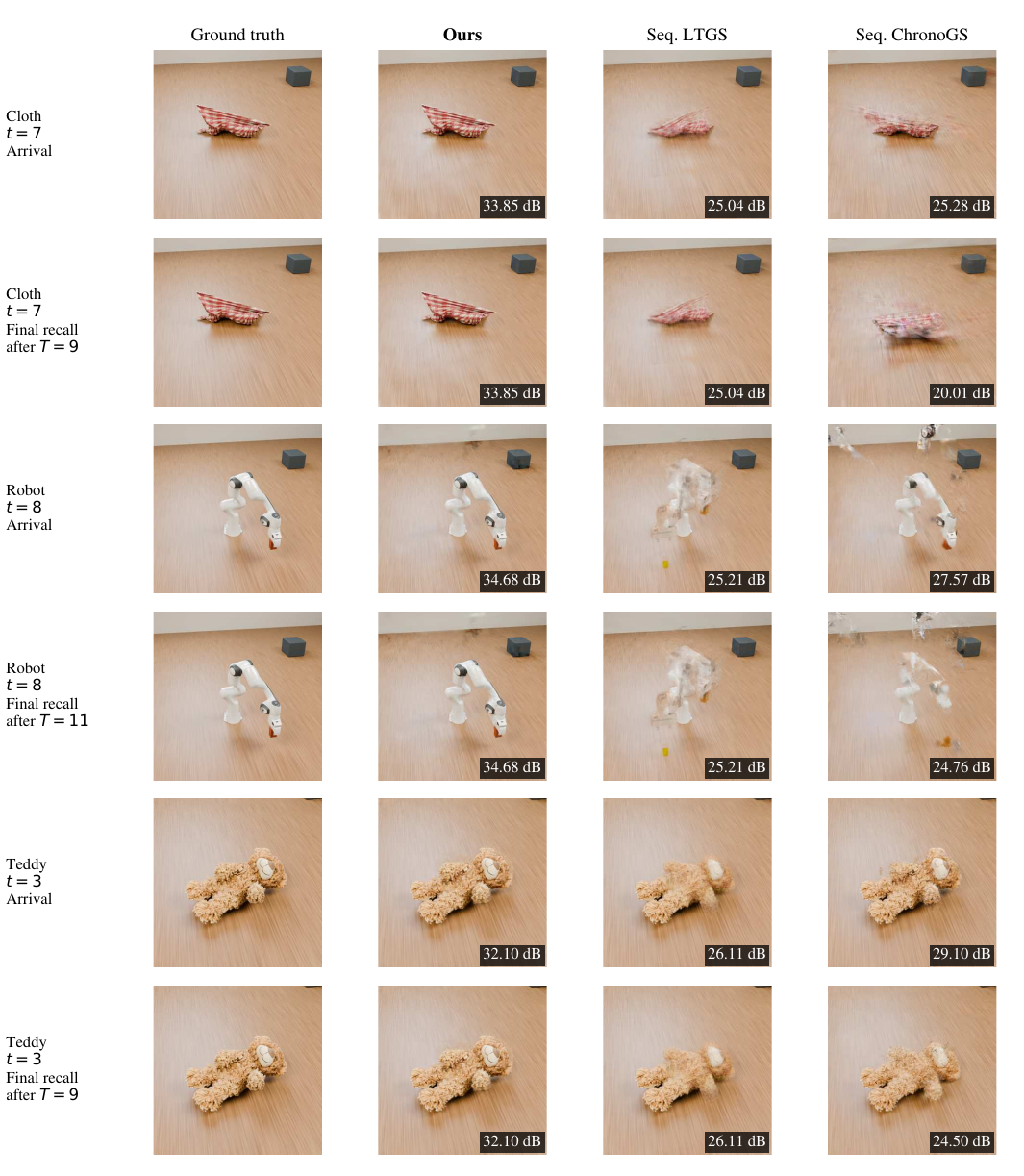}
\caption{\textbf{Arrival and final recall on synthetic Multi-State.} Cloth, Robot and Teddy. Each scene occupies two rows with identical ground truth and camera. Here, $t$ denotes the reconstructed visit and $T$ the final update. Overlays show whole-image PSNR (dB).}
\label{fig:supp-qual-synthetic}
\end{figure*}

\begin{figure*}[p]
\centering
\includegraphics[width=\textwidth,height=.86\textheight,keepaspectratio]{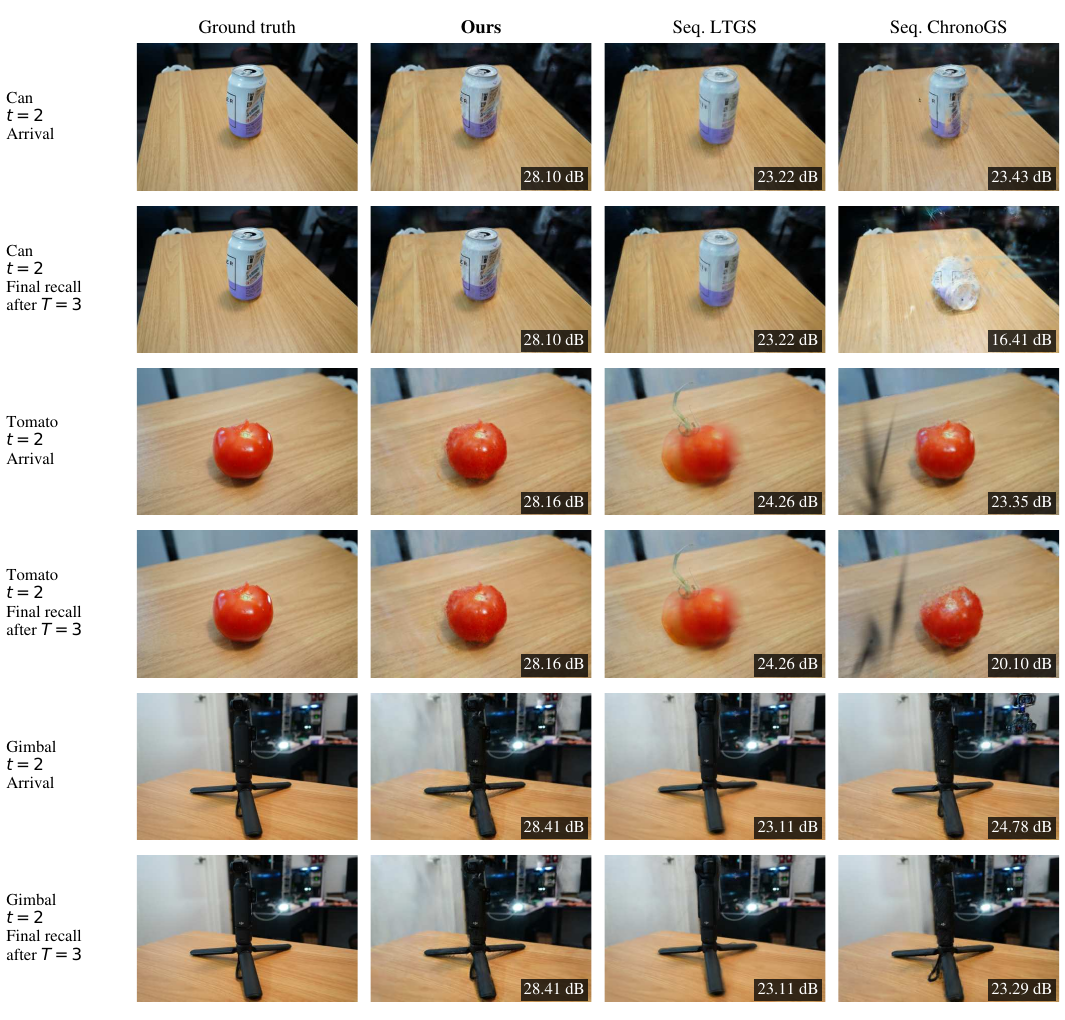}
\caption{\textbf{Arrival and final recall on real Multi-State.} Can, Tomato and Gimbal. Each scene occupies two rows with identical ground truth and camera. Here, $t$ denotes the reconstructed visit and $T$ the final update. Overlays show whole-image PSNR (dB).}
\label{fig:supp-qual-real}
\end{figure*}

\begin{figure*}[p]
\centering
\includegraphics[width=\textwidth,height=.86\textheight,keepaspectratio]{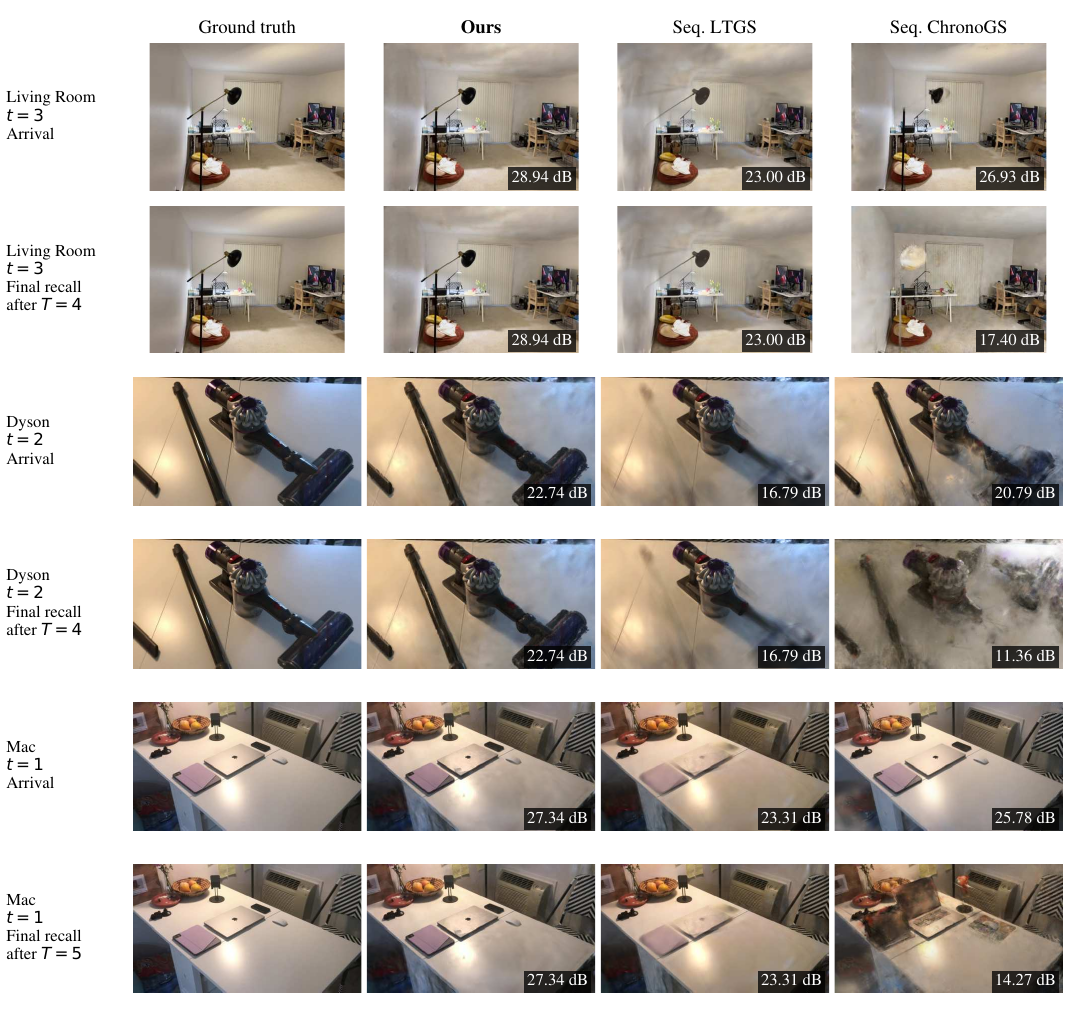}
\caption{\textbf{Arrival and final recall on WAT.} Living Room, Dyson and Mac. Each scene occupies two rows with identical ground truth and camera. Here, $t$ denotes the reconstructed visit and $T$ the final update. Overlays show whole-image PSNR (dB).}
\label{fig:supp-qual-wat}
\end{figure*}

\begin{figure*}[p]
\centering
\includegraphics[width=\textwidth,height=.86\textheight,keepaspectratio]{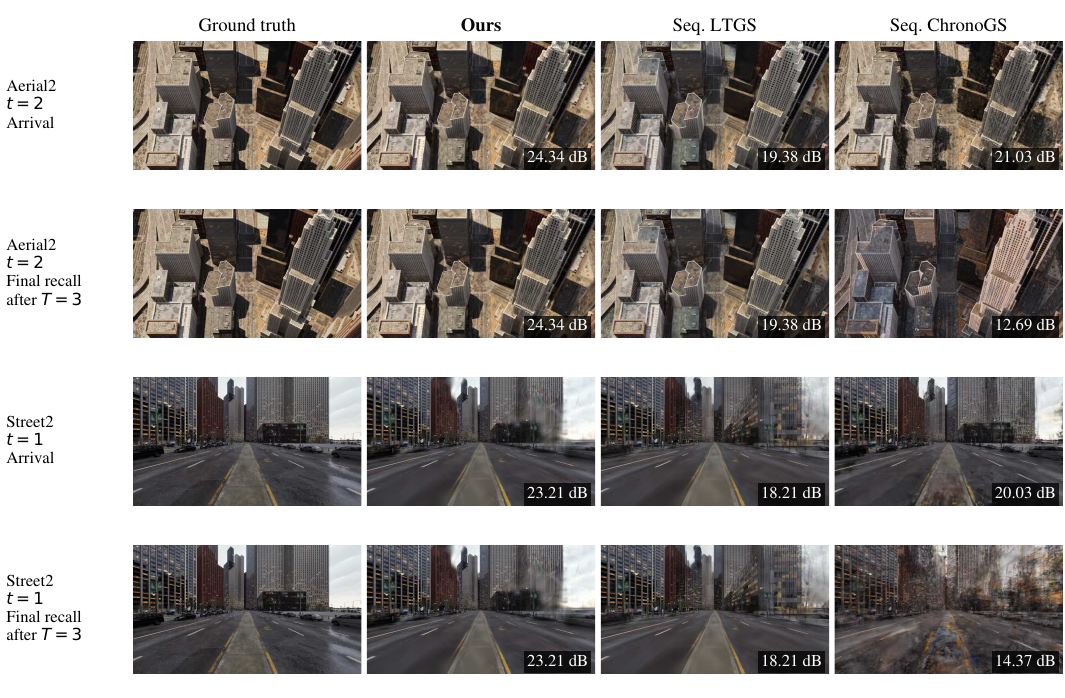}
\caption{\textbf{Arrival and final recall on ChronoScene.} Aerial2 and Street2. Each scene occupies two rows with identical ground truth and camera. Here, $t$ denotes the reconstructed visit and $T$ the final update. Overlays show whole-image PSNR (dB).}
\label{fig:supp-qual-chronoscene}
\end{figure*}

Figures~\ref{fig:supp-qual-synthetic}--\ref{fig:supp-qual-chronoscene} compare our method with the baselines on synthetic Multi-State, real Multi-State, WAT and ChronoScene, respectively. For each scene, one visit is rendered from the same held-out camera at arrival and at final recall, so that the two rows show both how a new visit is reconstructed and how well it is retained after all later updates. Overlays report whole-image PSNR.

In these figures, ReScene preserves object shape and detail more clearly than LTGS, whose arrival errors remain in its stored reconstructions. ChronoGS exhibits additional distortions at final recall, including altered object configurations in Multi-State, displaced or missing content in WAT, and changes to building appearance in ChronoScene. However, ReScene’s stored updates reproduce each arrival reconstruction exactly, including any errors it contains. Exact recovery therefore preserves an earlier state but does not improve it. For example, the background of Street2 in ChronoScene, which is blurred at arrival, remains equally blurred at final recall.

\begin{figure*}[t]
    \centering
    \includegraphics[width=\textwidth]{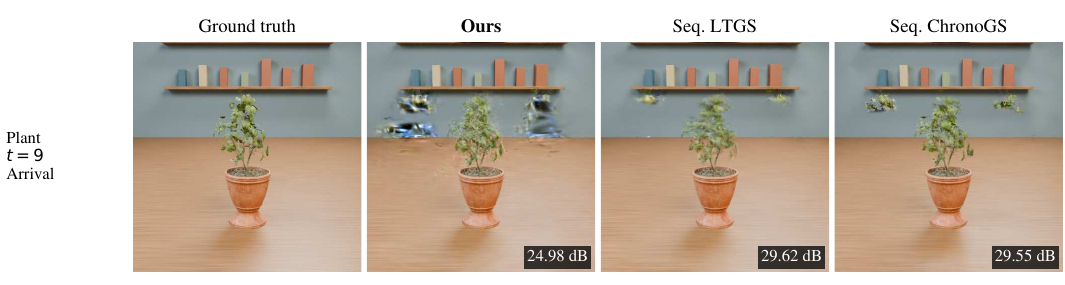}
    \caption{\textbf{Failure case on Plant.}
    Comparison at visit 9 from held-out view 0.
    Both sequential baselines achieve higher whole-image PSNR than
    our method in this example, illustrating a limitation when
    reconstructing thin structures with substantial occlusion.}
    \label{fig:plant-failure}
\end{figure*}

Meanwhile, figure~\ref{fig:plant-failure} illustrates a challenging case involving thin stems and leaves that occlude one another. Reconstructing these structures from sparse views remains difficult. All methods exhibit visible background artefacts, which are more pronounced in our reconstruction. Both sequential baselines achieve higher whole-image PSNR, but none reconstructs this scene without visible errors. 


\clearpage

\clearpage
\phantomsection
\addcontentsline{toc}{section}{References}
{\small\bibliographystyle{ieeenat_fullname}\bibliography{references}}

@inproceedings{neusc, author={Haotong Lin and Qianqian Wang and Ruojin Cai and Sida Peng and Hadar Averbuch-Elor and Xiaowei Zhou and Noah Snavely}, title={Neural Scene Chronology}, booktitle={CVPR}, year={2023}, pages={20752--20761}, url={https://openaccess.thecvf.com/content/CVPR2023/html/Lin_Neural_Scene_Chronology_CVPR_2023_paper.html}}

@inproceedings{4dgs, author={Guanjun Wu and Taoran Yi and Jiemin Fang and Lingxi Xie and Xiaopeng Zhang and Wei Wei and Wenyu Liu and Qi Tian and Xinggang Wang}, title={{4D Gaussian Splatting} for Real-Time Dynamic Scene Rendering}, booktitle={CVPR}, year={2024}, pages={20310--20320}}

@inproceedings{deformable, author={Ziyi Yang and Xinyu Gao and Wen Zhou and Shaohui Jiao and Yuqing Zhang and Xiaogang Jin}, title={Deformable {3D Gaussians} for High-Fidelity Monocular Dynamic Scene Reconstruction}, booktitle={CVPR}, year={2024}, url={https://openaccess.thecvf.com/content/CVPR2024/html/Yang_Deformable_3D_Gaussians_for_High-Fidelity_Monocular_Dynamic_Scene_Reconstruction_CVPR_2024_paper.html}}

@inproceedings{nerf, author={Ben Mildenhall and Pratul P. Srinivasan and Matthew Tancik and Jonathan T. Barron and Ravi Ramamoorthi and Ren Ng}, title={{NeRF}: Representing Scenes as Neural Radiance Fields for View Synthesis}, booktitle={ECCV}, year={2020}}

@article{3dgs, author={Bernhard Kerbl and Georgios Kopanas and Thomas Leimk{\"u}hler and George Drettakis}, title={{3D Gaussian Splatting} for Real-Time Radiance Field Rendering}, journal={ACM Transactions on Graphics}, volume={42}, number={4}, year={2023}}

@inproceedings{clnerf, author={Zhipeng Cai and Matthias M{\"u}ller}, title={{CLNeRF}: Continual Learning Meets {NeRF}}, booktitle={ICCV}, year={2023}}

@inproceedings{clsplats, author={Jan Ackermann and Jonas Kulhanek and Shengqu Cai and Haofei Xu and Marc Pollefeys and Gordon Wetzstein and Leonidas J. Guibas and Songyou Peng}, title={{CL-Splats}: Continual Learning of Gaussian Splatting with Local Optimization}, booktitle={ICCV}, year={2025}}

@inproceedings{gaussianupdate, author={Lin Zeng and Boming Zhao and Jiarui Hu and Xujie Shen and Ziqiang Dang and Hujun Bao and Zhaopeng Cui}, title={{GaussianUpdate}: Continual {3D Gaussian Splatting} Update for Changing Environments}, booktitle={ICCV}, year={2025}}

@article{ltgs, author={Minkwan Kim and Seungmin Lee and Junho Kim and Young Min Kim}, title={{LTGS}: Long-Term Gaussian Scene Chronology From Sparse View Updates}, journal={arXiv preprint arXiv:2510.09881}, year={2025}}

@article{chronogs, author={Zhongtao Wang and Jiaqi Dai and Qingtian Zhu and Yilong Li and Mai Su and Fei Zhu and Meng Gai and Shaorong Wang and Chengwei Pan and Yisong Chen and Guoping Wang}, title={{ChronoGS}: Disentangling Invariants and Changes in Multi-Period Scenes}, journal={arXiv preprint arXiv:2511.18794}, year={2025}}

@inproceedings{mast3r, author={Vincent Leroy and Yohann Cabon and J{\'e}r{\^o}me Revaud}, title={Grounding Image Matching in {3D} with {MASt3R}}, booktitle={ECCV}, year={2024}}

@inproceedings{clnerfwu, author={Xiuzhe Wu and Peng Dai and Weipeng Deng and Handi Chen and Yang Wu and Yan-Pei Cao and Ying Shan and Xiaojuan Qi}, title={{CL-NeRF}: Continual Learning of Neural Radiance Fields for Evolving Scene Representation}, booktitle={NeurIPS}, year={2023}}

@inproceedings{colmap,
  author    = {Sch{\"o}nberger, Johannes Lutz and Frahm, Jan-Michael},
  title     = {Structure-from-Motion Revisited},
  booktitle = {Proceedings of the IEEE Conference on Computer Vision
               and Pattern Recognition (CVPR)},
  pages     = {4104--4113},
  year      = {2016}
}

@inproceedings{nerfw,
  title = {{NeRF} in the Wild: Neural Radiance Fields for Unconstrained Photo Collections},
  author = {Ricardo Martin-Brualla and Noha Radwan and Mehdi S. M. Sajjadi and Jonathan T. Barron and Alexey Dosovitskiy and Daniel Duckworth},
  booktitle = {CVPR},
  year = {2021}
}

@inproceedings{wildgaussians,
  title = {{WildGaussians}: {3D} Gaussian Splatting in the Wild},
  author = {Jonas Kulhanek and Songyou Peng and Zuzana Kukelova and Marc Pollefeys and Torsten Sattler},
  booktitle = {NeurIPS},
  year = {2024}
}

@inproceedings{regnerf,
  title = {{RegNeRF}: Regularizing Neural Radiance Fields for View Synthesis from Sparse Inputs},
  author = {Michael Niemeyer and Jonathan T. Barron and Ben Mildenhall and Mehdi S. M. Sajjadi and Andreas Geiger and Noha Radwan},
  booktitle = {CVPR},
  year = {2022}
}

@inproceedings{fsgs,
  title = {{FSGS}: Real-Time Few-Shot View Synthesis Using Gaussian Splatting},
  author = {Zehao Zhu and Zhiwen Fan and Yifan Jiang and Zhangyang Wang},
  booktitle = {ECCV},
  year = {2024}
}

@inproceedings{dngaussian,
  title = {{DNGaussian}: Optimizing Sparse-View {3D} Gaussian Radiance Fields with Global-Local Depth Normalization},
  author = {Jiahe Li and Jiawei Zhang and Xiao Bai and Jin Zheng and Xin Ning and Jun Zhou and Lin Gu},
  booktitle = {CVPR},
  year = {2024}
}

@inproceedings{pixelsplat,
  title = {{pixelSplat}: {3D} Gaussian Splats from Image Pairs for Scalable Generalizable {3D} Reconstruction},
  author = {David Charatan and Sizhe Lester Li and Andrea Tagliasacchi and Vincent Sitzmann},
  booktitle = {CVPR},
  year = {2024}
}

@inproceedings{mvsplat,
  title = {{MVSplat}: Efficient {3D} Gaussian Splatting from Sparse Multi-View Images},
  author = {Yuedong Chen and Haofei Xu and Chuanxia Zheng and Bohan Zhuang and Marc Pollefeys and Andreas Geiger and Tat-Jen Cham and Jianfei Cai},
  booktitle = {ECCV},
  year = {2024}
}

@inproceedings{dust3r,
  title = {{DUSt3R}: Geometric {3D} Vision Made Easy},
  author = {Shuzhe Wang and Vincent Leroy and Yohann Cabon and Boris Chidlovskii and Jerome Revaud},
  booktitle = {CVPR},
  year = {2024}
}

@inproceedings{dnerf,
  title = {{D-NeRF}: Neural Radiance Fields for Dynamic Scenes},
  author = {Albert Pumarola and Enric Corona and Gerard Pons-Moll and Francesc Moreno-Noguer},
  booktitle = {CVPR},
  year = {2021}
}

@inproceedings{nerfies,
  title = {{Nerfies}: Deformable Neural Radiance Fields},
  author = {Keunhong Park and Utkarsh Sinha and Jonathan T. Barron and Sofien Bouaziz and Dan B. Goldman and Steven M. Seitz and Ricardo Martin-Brualla},
  booktitle = {ICCV},
  year = {2021}
}

@inproceedings{luiten2024dynamic,
  title = {Dynamic {3D} Gaussians: Tracking by Persistent Dynamic View Synthesis},
  author = {Jonathon Luiten and Georgios Kopanas and Bastian Leibe and Deva Ramanan},
  booktitle = {3DV},
  year = {2024}
}

@inproceedings{spacetimegs,
  title = {Spacetime Gaussian Feature Splatting for Real-Time Dynamic View Synthesis},
  author = {Zhan Li and Zhang Chen and Zhong Li and Yi Xu},
  booktitle = {CVPR},
  year = {2024}
}

@article{instantsplat,
  title = {{InstantSplat}: Sparse-View Gaussian Splatting in Seconds},
  author = {Zhiwen Fan and Wenyan Cong and Kairun Wen and Kevin Wang and Jian Zhang and Xinghao Ding and Danfei Xu and Boris Ivanovic and Marco Pavone and Georgios Pavlakos and Zhangyang Wang and Yue Wang},
  journal = {arXiv preprint arXiv:2403.20309},
  year = {2024},
  doi = {10.48550/arXiv.2403.20309}
}

@article{hypernerf,
  title = {{HyperNeRF}: A Higher-Dimensional Representation for Topologically Varying Neural Radiance Fields},
  author = {Keunhong Park and Utkarsh Sinha and Peter Hedman and Jonathan T. Barron and Sofien Bouaziz and Dan B. Goldman and Ricardo Martin-Brualla and Steven M. Seitz},
  journal = {ACM Transactions on Graphics},
  volume = {40}, number = {6}, pages = {238:1--238:12}, year = {2021},
  doi = {10.1145/3478513.3480487}
}
\end{document}